\documentclass[final,5p,times,twocolumn]{elsarticle}

\makeatletter
\def\ps@pprintTitle{%
 \let\@oddhead\@empty
 \let\@evenhead\@empty
 \def\@oddfoot{\hspace{9.1cm}{\thepage}}%
 \let\@evenfoot\@oddfoot}
\makeatother

\usepackage{amssymb}
\usepackage{amsthm}
\usepackage{amsfonts}

\usepackage{graphicx}
\usepackage{algorithm}
\usepackage{algorithmic}
\usepackage{accents}
\usepackage{multirow}
\usepackage{accents}

\usepackage{epstopdf}

\usepackage{array}
\usepackage{bm}

\usepackage{hyperref}
\usepackage{adjustbox}

\journal{Information Fusion}

\begin{document}

\begin{frontmatter}



\title{An experimental study of graph-based semi-supervised classification \\ with additional node information}

 \author[ucl]{B. Lebichot\corref{cor1}}
 \ead{bertrand.lebichot@uclouvain.be}
 \cortext[cor1]{Corresponding author}
 
 \author[ucl]{M. Saerens}
 \ead{bertrand.lebichot@uclouvain.be}
  
 \address[ucl]{Universite catholique de Louvain, ICTEAM \& LSM}


\author{}

\address{}

\begin{abstract}
The volume of data generated by internet and social networks is increasing every day, and there is a clear need for efficient ways of extracting useful information from them. As those data can take different forms, it is important to use all the available data representations for prediction. 
In this paper, we focus our attention on supervised classification using both regular plain, tabular, data and structural information coming from a network structure. 14 techniques are investigated and compared in this study and can be divided in three classes: the first one uses only the plain data to build a classification model, the second uses only the graph structure and the last uses both information sources. The relative performances in these three cases are investigated. Furthermore, the effect of using a graph embedding and well-known indicators in spatial statistics is also studied. 
Possible applications are automatic classification of web pages or other linked documents, of people in a social network or of proteins in a biological complex system, to name a few.
Based on our comparison, we draw some general conclusions and advices to tackle this particular classification task: some datasets can be better explained by their graph structure (graph-driven), or by their feature set (features-driven). The most efficient methods are discussed in both cases.

\end{abstract}


\end{frontmatter}





\section{Introduction}
\label{intro}

Nowadays, with the increasing volume of data generated, for instance by internet and social networks, there is a need for efficient ways to predict useful information from those data. Numerous data mining, machine learning and pattern recognition algorithms were developed for predicting information from a labeled database. These data can take several different forms and, in that case, it would be useful to use these alternative views in the prediction model. In this paper, we focus our attention on supervised classification using both regular tabular data and structural information coming from graphs or networks.

Many different approaches have been developed for information fusion in machine learning, pattern recognition and applied statistics, such as simple weighted averages (see, e.g., \cite{Cooke-1991}, \cite{Jacobs-1995}), Bayesian fusion (see, e.g., \cite{Cooke-1991}, \cite{Jacobs-1995}), majority vote (see, e.g., \cite{Chen-2001}, \cite{Kittler-2003}, \cite{Lad-1996}), models coming from uncertainty reasoning: fuzzy logic, possibility theory \cite{Klir-1988} (see, e.g., \cite{Dubois-1999}), standard multivariate statistical analysis techniques such as correspondence analysis \cite{Merz-1999}, maximum entropy modeling (see, e.g., \cite{Levy-1994}, \cite{Myung-1996}, \cite{FoussMaxEnt-2004}).

As well-known, the goal of classification is to automatically label data to predefined classes. This is also called supervised learning since it uses known labels (the desired prediction of an instance) to fit the classification model. One alternative is to use semi-supervised learning instead \cite{Zhu-2009,Chapelle-2006,Hill-2006,Macskassy-07}. 

Indeed, traditional pattern recognition, machine learning or data mining classification methods require large amounts of labeled training instances -- which is often difficult to obtain -- to fit accurate models. Semi-supervised learning methods can reduce the effort by including unlabeled samples. This name comes from the fact that the used dataset is a mixture of supervised and unsupervised data (it contains training samples that are unlabeled). Then, the classifier takes advantage from both the supervised and unsupervised data. The advantage here is that unlabeled data are often much less costly than labeled data. This technique allows to reduce the amount of labeled instances needed to achieve the same level of classification accuracy \cite{Zhu-2009,Chapelle-2006,Hill-2006,Macskassy-07}. In other words, exploiting the distribution of unlabeled data during the model fitting process can prove helpful.

Semi-supervised classification comes in two different settings: inductive and transductive \cite{Zhu-2009}. The goal of the former setting is to predict the labels of future test data, unknown when fitting the model, while the second is to classify (only) the unlabeled instances of the training sample. Some often-used semi-supervised algorithms include: expectation-maximization with generative mixture models, self-training, co-training, transductive support vector machines, and graph-based methods \cite{Daudin08,Joachims99,Zhu-2008}.

\begin{figure}[t!]
	\begin{center}
		\includegraphics[scale = 1.1]{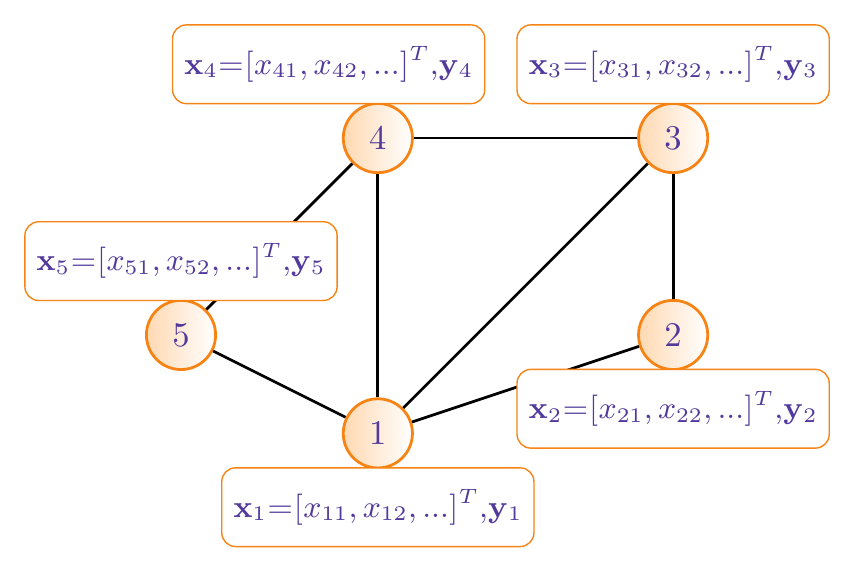}
		\caption{An example of graph with additional node information. Each node is characterized by a feature vector and a class label.}
		\label{FIG}
	\end{center}
	\end{figure}

The structure of the data can also be of different types. This paper focuses on a particular data structure: we assume that our dataset takes the form of a network with features associated to the nodes. Nodes are the samples of our dataset and links between these nodes represent a given type of relation between these samples. For each node, a number of features or attributes characterizing it is also available (see Figure \ref{FIG} for an example). Other data structures exist but are not studied in this paper; for instance:
\begin{itemize}
		\item Different types of nodes can be present, with different types of features sets describing them.
		\item Different types of relations can link the different nodes.
\end{itemize}

This problem has numerous applications such as classification of individuals in social networks, categorization of linked documents (e.g. patents or scientific papers), or protein function prediction, to name a few. In this kind of application (as in many others), unlabeled data are usually available in large quantities and are easy to collect: 
friendship links can be recorded on Facebook, text documents can be crawled from the internet and DNA sequences of proteins are readily available from gene databases. 

In this work, we investigate experimentally various models combining information on the nodes of a graph and the graph structure. Indeed, it has been shown that network information improves significantly prediction accuracy in a number of contexts \cite{Hill-2006,Macskassy-07}. Indeed, 14 classification algorithms using various combinations of data sources, mainly described in \cite{Fouss-2016}, are compared. The different considered algorithms are described in Section \ref{DimR}, \ref{FCE} and \ref{GBC} . A standard support vector machine (SVM) classifier is used as a baseline algorithm, but we also investigated the ridge logistic regression classifier. The results and conclusions obtained with this second model were similar to the SVM and are therefore not reported in this paper.

In short, the main questions investigated in this work are:
\begin{itemize}
		\item Does the combination of features on nodes and network structure works better than using the features only?
		\item Does the combination of features on nodes and network structure works better than using the graph structure only?
		\item Which classifier performs best on network structure alone, without considering features on nodes?
		\item Which classifier performs best when combining information, that is, using network structure with features on nodes?
\end{itemize}
Finally, this comparison leads to some general conclusions and advices when tackling classification problems on network data with node features.

In summary, this work has four main contributions:
\begin{itemize}
		\item The paper reviews different algorithms used for learning from both a graph structure and node features. Some algorithms are inductive while some others are transductive.
		\item An empirical comparison of those algorithms is performed on ten real world datasets.
		\item It investigates the effect of extracting features from the graph structure (and some well-known indicators in spatial statistics) in a classification context.
		\item Finally, this comparison is used to draw general conclusions and advices to tackle graph-based classification tasks.
\end{itemize}

The remaining of this paper is organized as follows. Section \ref{BgNot} provides some background and notation. Section \ref{RelW} investigates related work. Section \ref{Surv} introduces the investigated classification methods. Then, Section \ref{Exp} presents the experimental methodology and the results. Finally, Section \ref{CCL} concludes the paper.

\section{Background and notation}
\label{BgNot}

This section aims to introduce the necessary theoretical background and notation used in the paper. 
Consider a weighted, undirected, strongly connected, graph or network $G$ (with no self-loop) containing a set of $n$ vertices $\mathcal{V}$ (or nodes) and a set of edges $\mathcal{E}$ (or arcs, links). The $n\times n$ \textbf{adjacency matrix} of the graph, containing non-negative affinities between nodes, is denoted as $\mathbf{A}$, with elements $a_{ij} \ge 0$.

Moreover, to each edge between node $i$ and $j$ is associated a non-negative number $c_{ij}\ge 0$. This number represents the \textbf{immediate cost of transition} from node $i$ to $j$. If there is no link between $i$ and $j$, the cost is assumed to take a large value, denoted by $c_{ij} = \infty$. The \textbf{cost matrix} $\mathbf{C}$ is an $n\times n$ matrix containing the $c_{ij}$ as elements. Costs are set independently of the adjacency matrix -- they are quantifying the cost of a transition according to the problem at hand. Now, if there is no reason to introduce a cost, we simply set $c_{ij} = 1$ (paths are penalized by their length) or $c_{ij} = 1/a_{ij}$ (in this case, $a_{ij}$ is viewed as a conductance and $c_{ij}$ as a resistance) -- this last setting will be used in the experimental section.

We also introduce the \textbf{Laplacian matrix} $%
\mathbf{L}$ of the graph, defined in the usual manner: 
\begin{equation}
\mathbf{L}=\mathbf{D}-\mathbf{A}  \label{Eq_laplacian01}
\end{equation}where
$\mathbf{D} = \mathbf{Diag}(\mathbf{A}\mathbf{e})$ is
the diagonal (out)degree matrix of the graph $G$ containing the $a_{i \bullet} = \sum_{j=1}^{n} a_{ij}$ on its diagonal and $\mathbf{e}$ is a column vector full of ones.
One of the properties of $\mathbf{L}$ is that its eigenvalues provide useful information about the connectivity of the graph \cite{Chung-1997}. The smallest eigenvalue of $\mathbf{L}$ is always equals to $0$, and the second smallest one is equals to $0$ only if the graph is composed of at least two connected components. This last value is called the algebraic connectivity.

Moreover, a \textbf{natural random walk} on $G$ is defined in the standard way. In node $i$, the random walker chooses the next edge to follow according to transition probabilities
\begin{equation}
	p_{ij}=\frac{a_{ij}}{\sum_{j'=1}^{n} a_{ij'}}
	\label{Pref}
\end{equation}
representing the probability of jumping from node $i$ to node $j \in \mathcal{S}ucc(i)$, the set of successor nodes of $i$. The corresponding $n\times n$ \textbf{transition probabilities matrix} will be denoted as $\mathbf{P}$ and is stochastic. Thus, the random walker chooses to follow an edge with a likelihood proportional to the affinity (apart from the sum-to-one normalization), therefore favoring edges with a large affinity.

Moreover, we will consider that each of the nodes of $G$ has the same set of $m$ features, or attributes, with no missing values. The column vector $\mathbf{x}_i$ contains the values of the $m$ features of node $i$ and $x_{ij}$ states for the value of feature $j$ taken by node $i$. Moreover, $\mathbf{X}$ will refer to the $n \times m$ data matrix containing the $\mathbf{x}_{i}^{T}$ on its rows.

Finally we define $\mathbf{y}$ as the column vector containing the class labels of the nodes. Moreover, $\mathbf{y}^{c}$ is a binary vector indicating whether or not a node belongs to class number $c$. 

Recall that the purpose of the classification tasks will to predict the class of the unlabeled data (in a transductive setting), or to predict new test data (in an inductive setting), while knowing the values of the features $\mathbf{X}$ for all the nodes of $G$ and the class labels $\mathbf{y}^{c}$ on the \emph{labeled} nodes only for each $c$. Our baseline classifier based on features only will be a linear support vector machines (SVM).

\section{Some related work}
\label{RelW}

The 14 investigated models are presented in the next Section \ref{Surv}.
In addition to those models, other approaches exist. 

For example \cite{Belkin-2006,He-2010} use a standard ridge regression model complemented by a Laplacian regularization term, and has been called the Laplacian regularized least squares. This option was investigated but provided poor results compared to reported models (an is therefore not reported). Note that using a logistic ridge regression as the base classifier was also investigated in this work but results are not reported here for conciseness as it provided performances similar to SVMs.

Laplacian support vector machines (LapSVMs) extend the SVM classifier in order to take the structure of the network into account. It exploits both the information on the nodes and the graph structure in order to categorize the nodes through its Laplacian matrix (see Section \ref{BgNot}). To this end, \cite{Belkin-2006} proposed to add a graph Laplacian regularization term to the traditional SVM cost function in order to obtain a semi-supervised version of this model. A matlab toolbox for this model is available but provided poor results in terms of performance and tractability.

Chakrabarti et al. developed, in the context of patents classification \cite{Chakrabarti-1998}, a naive Bayes model in the presence of structural autocorrelation. The main idea is to use a naive Bayes classifier (see for example \cite{Bishop95,Hastie-2009,Theodoridis03}) combining both feature information on the nodes and structural information by making some independence assumptions. More precisely, it is assumed that the label of a node is influenced by two sources: features of the node and labels of the neighboring nodes (and does not depend on other information). It first considers that labels of all neighboring nodes are known and then relax this constrain by using a kind of relaxation labeling (see \cite{Chakrabarti-1998} for details). However, we found out that this procedure is very time consuming, even for small-size networks, and decided to not include it in the present work.

Other semi-supervised classifiers based on network data only (features on nodes are not available) were also developed \cite{Macskassy-07,Zhu-2008}. The interested reader is invited to read, e.g., \cite{MOI,Abney-2008,Chapelle-2006,Fouss-2016,Hofmann-2008,Silva-2016,Subramanya-2014,Zhu-2008,Zhu-2009b} (and included references), focused on this topic for a comprehensive description. Finally, an interesting survey and a comparative experiment of related methods can be found in \cite{Macskassy-07}. 

\section{Survey of relevant classification methods}
\label{Surv}
		
The different classification methods compared in this work are briefly presented in this section, which is largely inspired by \cite{Fouss-2016}. For a more thorough presentation, see the provided references to the original work or \cite{Fouss-2016}. The classification models are sorted into different families: graph embedding-based classifiers, extensions of feature-based classifiers, and graph-based classifiers.

\subsection{Graph embedding-based classifiers}
\label{DimR}

A first interesting way to combine information from the features on the nodes and from the graph structure is to perform a \emph{graph embedding} projecting the nodes of the graph into a low-dimensional space (an embedding space) preserving as much as possible its structural information, and then use the coordinates of the projected nodes as \emph{additional features} in a standard classification model, such as a logistic regression or a support vector machine.

This procedure has been proposed in the field of spatial statistics for ecological modeling \cite{Borcard-2002,Dray-2006,Meot-1993}, but also more recently in data mining \cite{Tang-2009,Tang-2009b,Tang-2010,Zhang-2008,Zhang-2008b}. While many graph embedding techniques could be used, \cite{Dray-2006} suggests to exploit Moran's or Geary's index of spatial autocorrelation in order to compute the embedding.

Let us briefly develop their approach (see \cite{Fouss-2016}). Moran's $I$ and Geary's $c$ (see, e.g., \cite{Haining-2003,Pfeiffer-2008,Waldhor-2006,Waller-2004}) are two coefficients commonly used in spatial statistics in order to test the hypothesis of spatial autocorrelation of a continuous measure defined on the nodes. Four possibilities will be investigated to extract features from the graph structure: maximizing Moran's $I$, minimizing Geary's $c$, local principal component analysis and maximizing the Bag-of-Path (BoP) modularity.

\subsubsection{Maximizing Moran's $I$}
\label{Moran}

Moran's $I$ is given by
\begin{equation}
I(\mathbf{x}) = \frac{n}{a_{\bullet \bullet}} \frac{\sum_{i,j=1}^n a_{ij} (x_i - \bar{x})(x_j - \bar{x})}{\sum_{i'=1}^n (x_{i'} - \bar{x})^2}
\label{Eq_Moran_s_index01}
\end{equation}
where $x_i$ and $x_j$ are the values observed on nodes $i$ and $j$ respectively, for a considered quantity measured on the nodes. The column vector $\mathbf{x}$ is the vector containing the values $x_i$ and $\bar{x}$ is the average value of $\mathbf{x}$. Then, $a_{\bullet \bullet}$ is simply the sum of all entries of $\mathbf{A}$ -- the volume of the graph.

$I(\mathbf{x})$ can be interpreted as a correlation coefficient similar to the Pearson correlation coefficient \cite{Haining-2003,Pfeiffer-2008,Waldhor-2006,Waller-2004}. The numerator is a measure of covariance among the neighboring $x_{i}$ in $G$, while the denominator is a measure of variance. $I$ is in the interval $[-1,+1]$. A value close to zero indicates no evidence of autocorrelation, a positive value indicates positive autocorrelation and a negative value indicates negative autocorrelation. Autocorrelation means that close nodes tend to take similar values. 

In matrix form, Equation (\ref{Eq_Moran_s_index01}) can be rewritten as
\begin{equation}
I(\mathbf{x}) = \frac{n}{a_{\bullet \bullet}} \frac{\mathbf{x}^{{T}} \mathbf{H} \mathbf{A} \mathbf{H} \mathbf{x}}{\mathbf{x}^{{T}} \mathbf{H} \mathbf{x}}
\label{Eq_Moran_s_index02}
\end{equation}
where $\mathbf{H} = (\mathbf{I} - \mathbf{E}/n)$ is the centering matrix \cite{Mardia-1978} and $\mathbf{E}$ is a matrix full of ones. Note that the centering matrix is idempotent, $\mathbf{H} \mathbf{H} = \mathbf{H}$.

The objective is now to find the score $\mathbf{x}$ that achieves the largest autocorrelation, as defined by Moran's index. This corresponds to the values that most explains the structure of $G$. It can be obtained by setting the gradient equal to zero; we then obtain the following generalized eigensystem:
\begin{equation}
\mathbf{H} \mathbf{A} \mathbf{H} \mathbf{x}' = \lambda \mathbf{x}' \textnormal{, and then } 
\mathbf{x} = \mathbf{H} \mathbf{x}'
\label{Eq_Moran_eigensystem02}
\end{equation}


The idea is thus to extract the first eigenvector $\mathbf{x}_{1}$ of the centered adjacency matrix (\ref{Eq_Moran_eigensystem02}) corresponding to the \emph{largest} eigenvalue $\lambda_1$ and then to compute the second-largest eigenvector, $\mathbf{x}_{2}$, orthogonal to $\mathbf{x}_{1}$, etc. The eigenvalues $\lambda_{i}$ are proportional to the corresponding Moran's $I(\mathbf{x}_{i})$.


The $p$ largest centered eigenvectors of (\ref{Eq_Moran_eigensystem02}) are thus extracted and then used as additional $p$ features for a supervised classification model (here an SVM). In other words, $\mathbf{X}_{\mathrm{Moran}} = [\mathbf{x}_{1},\mathbf{x}_{2}, \dots, \mathbf{x}_{p}]^{T}$ is a new data matrix, capturing the structural information of $G$, that can be concatenated to the feature-based data matrix $\mathbf{X}_{\mathrm{feature}}$, forming the extended data matrix $[\mathbf{X}_{\mathrm{feature}},\mathbf{X}_{\mathrm{Moran}}]$.

\subsubsection{Minimizing Geary's $c$}
\label{Geary}

On the other hand, Geary's $c$ is another weighted estimate of partial autocorrelation given by
\begin{equation}
c(\mathbf{x}) = \frac{(n-1)}{2 \, a_{\bullet \bullet}} \frac{\sum_{i,j=1}^n a_{ij} (x_i - x_j)^2}{\sum_{i'=1}^n (x_{i'} - \bar{x})^2} 
\label{Eq_Geary_s_index01}
\end{equation}
and is related to Moran's $I$. However, while Moran's $I$ considers a covariance between neighboring nodes, Geary's $c$ considers distances between pairs of neighboring nodes. It ranges from 0 to 2 with 0 indicating perfect positive autocorrelation and 2 indicating perfect negative autocorrelation \cite{Meot-1993,Pfeiffer-2008,Waller-2004}.

%

In matrix form, Geary's $c$ can be written as
\begin{equation}
c(\mathbf{x}) = \frac{(n-1)}{2 a_{\bullet \bullet}} \, \frac{\mathbf{x}^{{T}} \mathbf{L} \mathbf{x}}{\mathbf{x}^{{T}} \mathbf{H} \mathbf{x}}.
\label{Eq_Geary_s_index02}
\end{equation}

This time, the objective is to find the score vector minimizing Geary's $c$. By proceeding as for Moran's $I$, we find that minimizing $c(\mathbf{x})$ aims to compute the $p$ \emph{lowest} non-trivial eigenvectors of the Laplacian matrix:
\begin{equation}
\mathbf{L} \mathbf{x} = \lambda \mathbf{H} \mathbf{x}
\label{Eq_Geary_eigensystem02}
\end{equation}
and then use these eigenvectors as additional $p$ features in a classification model. We therefore end up with the problem of computing the lowest eigenvectors of the Laplacian matrix, which also appears in spectral clustering (ratio cut, see, e.g., \cite{Luxburg-2007,Fouss-2016,Newman-2010}). 

Geary's $c$ has a computational advantage over Moran's $I$: the Laplacian matrix is usually sparse, which is not the case for Moran's $I$. Moreover, note that since the Laplacian matrix $\textbf{L}$ is centered, any solution of $\mathbf{L} \mathbf{x} = \lambda \mathbf{x}$ is also a solution of Equation (\ref{Eq_Geary_eigensystem02}).

\subsubsection{Local principal component analysis}           
\label{LPCA}

In \cite{Benali-1990,Lebart-2000}, the authors propose to use a measure of local, structural, association between nodes. The \textbf{contiguity ratio} \index{Contiguity ratio} is defined as
\begin{equation} 
cr(\mathbf{x}) =  \frac{ \sum_{i=1}^n (x_i - m_i)^2 }{ {\sum_{i'=1}^n (x_{i'} - \bar{x})^2} } {,} \label{Eq131}
 \textnormal{ with } m_i = \sum_{j \in \mathcal{N}(i)} p_{ij} x_{j}.  
\label{Eq132}
\end{equation}
and $m_i$ is the average value observed on the neighbors of $i$, $\mathcal{N}(i)$. As for Geary's index, the value is close to zero when there is a strong structural association. However, there are no clear bounds indicating no structural association or negative correlation \cite{Lebart-2000}.

The numerator of Equation (\ref{Eq131}) is the mean squared difference between the value on a node and the average of its neighboring values; it is called the local variance in \cite{Lebart-2000}. The denominator is the standard sample variance. In matrix form,
\begin{equation}
cr(\mathbf{x}) 
= \frac{ \mathbf{x}^{{T}} (\mathbf{I} - \mathbf{P})^{{T}} (\mathbf{I} - \mathbf{P}) \mathbf{x} }{ \mathbf{x}^{{T}} \mathbf{H} \mathbf{x} }.
\end{equation}

Proceeding as for Geary and Moran's indexes, minimizing $cr(\mathbf{x})$ aims to solve
\begin{equation}
(\mathbf{I} - \mathbf{P})^{{T}} (\mathbf{I} - \mathbf{P}) \mathbf{x} = \lambda \mathbf{H} \mathbf{x}
\label{Eq_local_PCA_eigensystem02}
\end{equation}
Here again, eigenvectors corresponding to the smallest non-trivial eigenvalues of the eigensystem (\ref{Eq_local_PCA_eigensystem02}) are extracted. This procedure is also referred to as local principal component analysis in \cite{Lebart-2000}.

\subsubsection{Bag-of-path modularity}
\label{BoPMod}
		For this algorithm, we also compute a number of structural features, but now derived from the modularity measure computed in the bag-of-path (BoP) framework \cite{Robin-2014}, and concatenate them to the node features $[\mathbf{X}_{\mathrm{feature}},{\mathbf{X}_{\mathrm{BoPMod}}}]$. Again, a SVM is then used to classify all unlabeled nodes. Indeed, it has been shown that using the dominant eigenvectors of the BoP modularity matrix provides better performances than using the eigenvectors of the standard modularity matrix . The result for the standard modularity matrix are therefore not reported here.
		
		It can be shown (see \cite{Robin-2014} for details) that the BoP modularity matrix is equal to
			\begin{equation}	
				\mathbf{Q}_\mathrm{BoP} = \mathbf{Z} - \frac{(\mathbf{Z} \mathbf{e}) (\mathbf{e}^{T}\mathbf{Z})}{\mathbf{e}^{T} \mathbf{Z} \mathbf{e}}
			\end{equation}
			where $\mathbf{Z}$ is the fundamental bag-of-path $n \times n$ matrix and $\mathbf{e}$ is a length $n$ column vector full of ones. Then as for Moran's $I$ and Geary's $c$, an eigensystem
			\begin{equation}	
				\mathbf{Q}_\mathrm{BoP} \mathbf{x} = \lambda \mathbf{x}
			\end{equation}
must be solved and the largest eigenvectors are used as new, additional, structural, features.

\subsection{Extensions of standard feature-based classifiers}
\label{FCE}

These techniques rely on extensions of standard feature-based classifiers (for instance a logistic regression model or a support vector machine). The extension is defined in order to take the network structure into account. 

\subsubsection{The AutoSVM: taking autocovariates into account}
\label{auto}

This model is also known as the \textbf{autologistic} 
\index{Autologistic model} or \textbf{autologit} model \index{Autologit model} \cite{Bezag-1972,Augustin-1996,Augustin-1998,Lu-2003}, and is frequently used in the spatial statistics and biostatistics fields. 

Note that, as a SVM is used as base classifier in this work (see Section \ref{intro}), we adapted this model (instead of the logistic regression in \cite{Augustin-1998}) in order to take the graph structure into account.
The method is based on the quantity $ac_{i}^{c} = \sum_{j \in \mathcal{N}(i)} p_{ij} \hat{y}^{c}_{j}$, where $\hat{y}^{c}_{j}$ is the predicted membership of node $j$, called the \textbf{autocovariate} in \cite{Augustin-1998} (other forms are possible, see \cite{Augustin-1996,Augustin-1998}). It corresponds to the weighted averaged membership to class $c$ within the neighborhood of $i$: it indicates to which extent neighbors of $i$ belong to class $c$. The assumption is that node $i$ has a higher chance to belong to class $c$ if its neighbors also belong to that class. $\mathbf{Ac}$ will be the matrix containing $ac_{i}^{c}$ for all $i$ and $c$.


However, since the predicted value $\hat{y}^{c}_{j}$ depends on the occurrence of the predicted value on other nodes, building the model is not straightforward. For the autologistic model, it goes through the maximization of the (pseudo-)likelihood (see for example \cite{Pawitan-2001,Bezag-1972}), but we will consider another alternative \cite{Augustin-1998} which uses a kind of expectation-maximization-like heuristics (EM, see, e.g. \cite{Dempster77,Mclachlan-2008}), and is easy to adapt to our SVM classifier.

Here is a summary (see \cite{Fouss-2016}) of the estimation procedure proposed in \cite{Augustin-1998}:
\begin{enumerate}
  \item At $t=0$, initialize the predicted class memberships $\hat{y}_{i}^{c}(t=0)$ of the unlabeled nodes by a standard SVM depending on the feature vectors only, from which we disregard the structural information (the information about neighbors' labels). For the labeled nodes, the membership values are not modified and are thus set to the true, observed, memberships.
  \item Compute the current values of the autocovariates, $ac_{i}^{c} = \sum_{j \in \mathcal{N}(i)} p_{ij} \hat{y}^{c}_{j}(t)$, for all nodes.
  \item Train a so-called \textbf{autoSVM} model based on these current autocovariate values as well as the features on nodes.
    This provides parameter estimates $\hat{\mathbf{w}}^{c}$.
  \item Compute the predicted class memberships $\hat{y}^{c}_{i}(t+1)$ of the set of unlabeled nodes from the fitted autoSVM model. This is done by sequentially selecting each unlabeled node $i$ in turn, and applying the fitted autoSVM model, obtained in step 3 based on the autocovariates of step 2. After having considered all the nodes, we have the new predicted values $\hat{y}^{c}_{i}(t+1)$.
  \item Steps 2 to 4 are be iterated until convergence of the predicted membership values $\hat{y}_{i}^{c}(t)$.
\end{enumerate}

\subsubsection{Double kernel SVM}
\label{DK SVM}

Here, we describe another simple way of combining the information coming from features on nodes and graph structure. The basic idea (\cite{Roth-2001,Fouss-2016}) is to
\begin{enumerate}
  \item Compute a $n \times n$ kernel matrix based on node features \cite{Scholkopf-2002,Shawe-Taylor-2004}, for instance a linear kernel or a gaussian kernel.
  \item Compute a $n \times n$ kernel matrix on the graph \cite{FoussKernelNN-2012,Fouss-2016,Gaertner-2008,Shawe-Taylor-2004}, for instance the regularized commute-time kernel (see Subsection \ref{Ugo}).
  \item Fit a SVM based on these two combined kernels.
\end{enumerate}
Then, by using the kernel trick, everything happens as if the new data matrix is
\begin{equation}
\mathbf{X}_{\mathrm{new}} = [\mathbf{K}_{\mathbf{A}}, \mathbf{K}_{\mathbf{X}}]
\end{equation}
where $\mathbf{K}_{G}$ is a kernel on a graph and $\mathbf{K}_{\mathbf{X}} = \mathbf{X} \mathbf{X}^{{T}}$
is the kernel matrix associated to the features on the nodes (see \cite{Fouss-2016} for details). Then, we can fit a SVM classifier based on this new data matrix and the class labels of labeled nodes.

\subsubsection{A spatial autoregressive model}
\label{SAR}

		This model is a spatial extension of a standard regression model \cite{LeSage-2009} and is well known in spatial econometrics. This extended model assumes that the vector of class memberships $\mathbf{y}^{c}$ is generated in each class $c$ according to
		\begin{equation}
			\mathbf{y}^{c}  = \rho \mathbf{P} \mathbf{y}^{c} + \mathbf{X} \mathbf{w}^{c} + \bm{\epsilon}
		\end{equation}
where $\mathbf{w}^{c}$ is the usual parameter vector, $\rho$ is a scalar parameter introduced to account for the structural dependency through $\mathbf{P}$ and $\epsilon$ is an error term. Obviously if $\rho$ is equal to zero, there is no structural dependency and the model reduce to a standard linear regression model. Lesage's Econometrics Matlab toolbox was used for the implementation of this model \cite{LeSage-2009}; see this reference for more information.

\subsection{Graph-based classifiers}
\label{GBC}

We also investigate some semi-supervised methods based on the graph structure only (no node feature exists or features are simply not taken into account). We selected the techniques performing best in a series of experimental comparisons \cite{FoussKernelNN-2012,MOI,Mantrach-2011}.
They rely on some strong assumptions about the distribution of labels: that neighboring nodes (or ``close nodes'') are likely to share the same class label \cite{Chapelle-2006}.

\subsubsection{The bag-of-paths group betweenness}
\label{BOP}
	
This model \cite{MOI} considers a bag containing all the possible paths between pairs of nodes in $G$. Then, a Boltzmann distribution, depending on a temperature parameter $T$, is defined on the set of paths such that long (high-cost) paths have a low probability of being picked from the bag, while short (low-cost) paths have a high probability of being picked. The \textbf{Bag-of-Paths (BoP) probabilities}, ${P}(s=i,e=j)$, providing the probability of drawing a path starting in $i$ and ending in $j$, can be computed in closed form and a betweenness measure quantifying to which extend a node is in-between two nodes is defined. A node receives a high betweenness if it has a large probability of appearing on paths connecting two arbitrary nodes of the network. A group betweenness between classes is defined as the sum of the contribution of all paths starting and ending in a particular class, and passing through the considered node. Each unlabeled nodes is then classified according to the class showing the highest group betweenness. More information can be found in \cite{MOI}.
	
	\subsubsection{A sum-of-similarities based on the regularized commute time kernel}
	\label{RCTK}
	
We finally investigate a classification procedure based on a simple alignment with the regularized commute time kernel $\mathbf{K}$, a \textbf{sum-of-similarities} defined by $\mathbf{K}\mathbf{y}^{c}$, with $\mathbf{K}=(\textbf{D}-\alpha \textbf{A})^{-1}$ \cite{Zhou-2003,FoussKernelNN-2012,Fouss-2016}. This expression quantifies to which extend each node is close (in terms of the similarity provided by the regularized commute time kernel) to class $c$. This similarity is computed for each class $c$ in turn. Then, each node is assigned to the class showing the largest sum of similarities. Element $i,j$ of this kernel can be interpreted as the discounted cumulated probability of visiting node $j$ when starting from node $i$. The (scalar) parameter $\alpha \in \left] 0,1 \right]$ corresponds to a killed random walk where the random walker has a $(1-\alpha$) probability of disappearing at each step. This method provided good results in a comparative study on graph-based semi-supervised classification \cite{FoussKernelNN-2012,Mantrach-2011,KEVIN}.

\section{Experiments}
\label{Exp}

In this section, the different classification methods will be compared on semi-supervised classification tasks and several datasets.
The goal is to classify unlabeled nodes in partially labeled graphs and to compare the results obtained by the different methods in terms of classification accuracy.

This section is organized as follows. First, the datasets used for semi-supervised classification are described in Subsection \ref{Data}. Then, the compared methods are recalled in Subsection \ref{Classif}. The experimental methodology is explained in Subsection \ref{ExpM}. Finally, results are presented and discussed in Subsection \ref{ResDis}.

\subsection{Datasets}
\label{Data}

\begin{table}[t]
\caption{Class distribution of the four \emph{WebKB cocite} datasets.}
\scriptsize
\begin{center}
\begin{tabular}{lcccc}\hline
 & \textbf{Cornell}& \textbf{Texas}& \textbf{Washington}& \textbf{Wisconsin}\\
\textbf{Class} & (DB1)& (DB2)& (DB3)& (DB4)\\
\hline
\\
Course		&42	&33	&59 &70	\\
Faculty		&32	&30	&25	&32	\\
Student		&83 &101&103&118\\
Project + staff		&38	&19	&28	&31	\\
\\
\textbf{Total} &195&183&230&251\\
\textbf{Majority class (\%)} &42.6&55.2&44.8&47.0\\
\\
\textbf{Number of features} &1704&1704&1704&1704\\
\hline
\end{tabular}
\label{tab: Tabcocite}
\end{center}
\end{table}

\begin{table}[t]
\caption{Class distribution of the three \emph{Ego facebook} datasets.}
\scriptsize
\begin{center}
\begin{tabular}{lcccc}\hline
 & \textbf{FB 107}& \textbf{FB 1684}& \textbf{FB 1912}\\
\textbf{Class} & (DB5)& (DB6)& (DB7)\\
\hline
\\
Main group			&737	&568	&524	\\
Other groups 		&308	&225	&232	\\
\\
\textbf{Total} &1045&793&756\\
\textbf{Majority class (\%)} &70.5&71.2&69.3\\
\\
\textbf{Number of features} & 576 & 319 & 480 \\
\hline
\end{tabular}
\label{tab: Tabfacebook}
\end{center}
\end{table}

\begin{table}[t]
\caption{Class distribution of the \emph{Citeseer, Cora} and \emph{Wikipedia} datasets.}
\scriptsize
\begin{center}
\begin{tabular}{lcccc}\hline
 & \textbf{Citeseer}& \textbf{Cora}& \textbf{Wikipedia}\\
\textbf{Class} & (DB8)& (DB9)& (DB10)\\
\hline
\\
Class 1			&269	&285	&248 \\
Class 2			&455	&406	&509	\\
Class 3			&300 	&726	&194\\
Class 4			&75		&379	&99	\\
Class 5			&78		&214	&152	\\
Class 6			&188	&131	&409	\\
Class 7			& 		&344	&181	\\
Class 8			& 		& 		&128	\\
Class 9			& 		& 		&364	\\
Class 10		& 		& 		&351	\\
Class 11		& 		& 		&194	\\
Class 12		& 		& 		&81	\\
Class 13		& 		& 		&233	\\
Class 14		& 		& 		&111	\\
\\
\textbf{Total} &1392&2708&3271\\
\textbf{Majority class (\%)} &32.7&26.8&15.6\\
\\
\textbf{Number of features} &3703&1434&4973\\
\hline
\end{tabular}
\label{tabothers}
\end{center}
\end{table}

All datasets are described by (i) the adjacency matrix $\mathbf{A}$ of the underlying graph, (ii) a class vector (to predict) and (iii) a number of features on nodes gathered in the data matrix $\mathbf{X}_{\mathrm{features}}$. Using a chi-square test, we kept only the 100 most significant variables for each dataset. The datasets are available at \url{http://www.isys.ucl.ac.be/staff/lebichot/research.htm}.

For each of these dataset, if more than one graph connected component is present, we only use the biggest connected component, deleting all the others nodes, features and target classes. Also, we choose to work with undirected graphs for all datasets: if a graph is directed, we used $\mathbf{A} = (\mathbf{A}^T + \mathbf{A})/2$ to introduce reciprocal edges.

\begin{itemize}
  \item \emph{WebKB cocite} (\textbf{DB1-DB4}) \cite{senaimag08}. These four datasets consist of web pages gathered from computer science departments from four universities (there is four datasets, one for each university), with each page manually labeled into one of four categories: course, faculty, student and project \cite{Macskassy-07}. The pages are linked by citations (if $x$ links to $y$ then it means that $y$ is cited by $x$). Each web page in the dataset is also characterized by a binary word vector indicating the absence/presence of the corresponding word from the dictionary. The dictionary consists of 1703 unique words (words appearing less than 10 times were ignored).
Originally, a fifth category, Staff, was present but since it contained only very few instances, it was merged with the Project class. Details on these datasets are shown in Table \ref{tab: Tabcocite}.
 


  \item The three \emph{Ego Facebook} datasets (\textbf{DB5-DB7}) \cite{McAuley2012} consist of circles (or friends communities) from Facebook. Facebook data were collected from survey participants using a Facebook application. The original dataset includes node features (profiles), circles, and ego networks for 10 networks. Those data are anonymized. We keep the three first networks and we define the classification task as follow: we picked the majority circle (the target circle) and aggregated all the others (non-target circles). 
    Details on these datasets are shown in Table \ref{tab: Tabfacebook}. Each dataset has two classes.
  	%

  \item The CiteSeer dataset  (\textbf{DB8}) \cite{senaimag08} consists of 3312 scientific publications classified into six classes. The pages are linked by citation (if $x$ links to $y$ then it means that $y$ is cited by $x$). Each publication in the dataset is described by a binary word vector indicating the absence/presence of the corresponding word from the dictionary. The dictionary consists of 3703 unique words (words appearing less than 10 times were ignored). Target variable is the domain of the publications (six topics, not reported here). Details on this dataset are shown in Table \ref{tabothers}.   
  

  \item The Cora dataset (\textbf{DB9}) \cite{senaimag08} consists of 2708 scientific publications classified into one of seven classes denoting topics as for previous dataset. Pages are linked by citations (if $x$ links to $y$ then it means that $y$ is cited by $x$). Each publication is also described by a binary word vector indicating the absence/presence of the corresponding word from the dictionary. The dictionary consists of 1434 unique words or features (words appearing less than 10 times were ignored). Target variable is the topic of the publications. Details on this dataset are shown in Table \ref{tabothers}.
  

  \item The Wikipedia dataset (\textbf{DB10}) \cite{senaimag08} consists of 3271 Wikipedia articles that appeared in the featured list in the period Oct.\ 7-21, 2009. Each document belongs to one of 14 distinct categories (such as Science, Sport, Art, \dots), which were obtained by using the category under which each article is listed. After stemming and stop-word removal, the content of each document is represented by a tf/idf-weighted feature vector, for a total of 4973 words. Pages are linked by citation (if $x$ links to $y$ then it means that $y$ is cited by $x$). Target variable is the articles field (14 different, not reported here).  Details on this dataset are shown in Table \ref{tabothers}.
  
  
\end{itemize}

Moreover, in order to study the impact of the relative information provided by the \emph{graph structure} and the \emph{features on nodes}, we created new derived datasets by \emph{weakening gradually} the information provided by the node features.
More precisely, for each dataset, the features available on the nodes have been ranked by decreasing association (using a chi-square statistics) with the target classes to be predicted. Then, datasets with \emph{subsets} of the features containing respectively the 5 (5F), 10 (10F), 25 (25F), 50 (50F) and 100 (100F) most informative features were created (sets of features). These datasets are weakened versions of the original datasets, allowing to investigate the respective role of features on nodes and graph structure. We also investigate sets with more features (200 and 400). Conclusions were the same so that they are not reported here for conciseness.

\subsection{Compared classification models} 
\label{Classif}

In this work, a transductive scheme is used, as we need to know the whole graph structure to label unlabeled nodes. 14 different algorithms will be compared and can be sorted in three categories, according to the information they use. Some algorithms use only features to build the model (denoted as X -- data matrix with features only), others use only the graph structure (denoted as A -- adjacency matrix of the graph only), and the third category uses both the structure of the graph and the features of the nodes (denoted as AX).

\subsubsection{Using features on nodes only}

\begin{itemize}
	\item[] This reduces to a standard classification problem and we use a linear Support Vector Machine (SVM) based on the features of the nodes to label these nodes (\textbf{SVM-X}). 
Here, we consider SVMs in the binary classification setting (i.e. $\mathbf{y}_i \in \{ -1, +1 \}$). For multiclass problems, we used a one-vs-one strategy \cite{Hsu2002}.
This classifier will be used as a baseline. In practical terms, we use the well-known Liblinear library \cite{REF08a}.
Notice that SVM follows an inductive scheme, unlike all other methods. Transductive SVMs \cite{Vapnik-1998} were also considered, but their implementation was too time-consuming to be included in the present analysis. 

\end{itemize}

\subsubsection{Using graph structure only}
\label{Ugo}

Three different families of methods using graph structure only are investigated.
\begin{itemize}
	\item The Bag of Path classifier based on the bag-of-paths group betweenness (\textbf{BoP-A}). This betweenness is computed for each class in turn. Then, each unlabeled node is assigned to the class showing the largest value (see Section \ref{BOP} for more details).
	\item The sum-of-similarities method based on the Regularized Commute Time Kernel (\textbf{CTK-A}).
	The classification procedure is the same as BoP-A: the class similarity is computed for each class in turn and each unlabeled node is assigned to the class showing the largest similarity (see Section \ref{RCTK} for more details).
	\item The four graph embedding techniques discussed in Section \ref{DimR} are used together with an SVM, without considering any node feature. The SVM is trained using a given number of extracted eigenvectors derived from each measure (this number is a parameter to tune). The SVM model is then used to classify the unlabeled nodes.
	\begin{itemize}
		\item SVM using Moran's $I$ derived dominant eigenvectors; see Section \ref{Moran} (\textbf{SVM-M-A}).
	
		\item SVM using Geary's $c$ derived dominant eigenvectors; see Section \ref{Geary} (\textbf{SVM-G-A}).

		\item SVM using the dominant eigenvectors extracted from local principal component analysis; see Section \ref{LPCA} (\textbf{SVM-L-A}).
		
		\item SVM using the dominant eigenvectors extracted from the bag-of-paths modularity; see Section \ref{BoPMod} (\textbf{SVM-BoPM-A}).

	\end{itemize}
\end{itemize}

\subsubsection{Using both information (features on nodes and graph structure)}

Here, we investigate:

\begin{itemize}

		\item A double kernel SVM (\textbf{DK SVM}).
		In this case, two kernels are computed, one defined on the graph and the second from the node features $\mathbf{X}_{\mathrm{new}} = [\mathbf{K_{\mathrm{A}}},\mathbf{K_{\mathrm{X}}]}$ (see Section \ref{DK SVM}). A SVM is then used to classify the unlabeled nodes.
		
		\item Support vector machine using Autocovariates (\textbf{ASVM-AX}).
		In this algorithm, autocovariates are added to the node features $\mathbf{X}_{\mathrm{new}} = [\mathbf{X}_{\mathrm{feat}},\mathbf{Ac}]$ (see Section \ref{auto}).

		\item Spatial AutoRegressive model (\textbf{SAR-AX}).
		This model is a spatial extension of the standard regression model (see Section \ref{SAR}), used to classify the unlabeled nodes.

\item The dominant eigenvectors (this number is a parameter to tune) provided by the four graph embedding techniques (Section \ref{DimR}) are combined with the node features and then injected into a linear SVM classifier:
	\begin{itemize}
	
		\item SVM using Moran's $I$-derived features (see \ref{Moran}) in addition to node features (\textbf{SVM-M-AX}): $\mathbf{X}_{\mathrm{new}} = [\mathbf{X}_{\mathrm{feat}},\mathbf{X}_{\mathrm{Moran}}]$.
	
		\item SVM using Geary's $c$-derived features (see \ref{Geary}) in addition to node features (\textbf{SVM-G-AX}): $\mathbf{X}_{\mathrm{new}} = [\mathbf{X}_{\mathrm{feat}},\mathbf{X}_{\mathrm{Geary}}]$.
	
		\item SVM using graph local principal component analysis (see \ref{LPCA}) in addition to node features (\textbf{SVM-L-AX}): $\mathbf{X}_{\mathrm{new}} = [\mathbf{X}_{\mathrm{feat}},\mathbf{X}_{\mathrm{LPC}}]$.
				
		\item SVM using bag-of-path modularity  (see \ref{BoPMod}) in addition to node features (\textbf{SVM-BoPM-AX}): $\mathbf{X}_{\mathrm{new}} = [\mathbf{X}_{\mathrm{feat}},\mathbf{X}_{\mathrm{BoP Mod}}]$.

	\end{itemize}
	
\end{itemize}

The considered classifiers, together with their parameters to be tuned, are listed in Table \ref{Param}.

\begin{table*} [htp] 

\caption{The 14 classifiers, the value range tested for tuning their parameters and the most frequently selected values: Mode is the most selected value across all datasets. Note that $p$, the number of extracted eigenvector, is given in $\%$: this is relative number of created features with respect to the number of node of the graph (different for each dataset).}
\small
\begin{adjustbox}{width=1\textwidth,center}
\begin{tabular}{|l|c|c|c|c|c|c|}
\hline
\textbf{Classification model} &\textbf{Use A} &\textbf{Use X} &  \textbf{Acronym} & \textbf{Param.} & \textbf{Tested values} & \textbf{Mode}\\ 
\hline
Bag of paths betweenness (\ref{BOP})													& yes	& no	& BoP-A     &  $\theta > 0					$ & $10^{[-9,-6,-3,0]}$ 								& $10^{-6} (40.2\%) $  \\
Sum of similarities based on the RCT kernel (\ref{RCTK})			& yes	& no	& CTK-A     &	 $\lambda > 0				  $ & $0.2,0.4,0.6,0.8,1$ 						 		& $0.8 (39.4\%) $  \\
SVM based on Moran's extracted features only (\ref{Moran})   								& yes	& no	&SVM-M-A   	&  $ C > 0							$ & $10^{[-6,-4,-2,0,2,4,6]}$						& $10^{-2} (63.0\%) $  \\
                                    &     &     &           &  $ p > 0							$ & $[5,10,20,35,50 \%]$              	& $5\% (74.0\%) $  \\ 
SVM based on Geary's extracted features only (\ref{Geary})										& yes	& no	&SVM-G-A   	&  $ C > 0							$ &	$10^{[-6,-4,-2,0,2,4,6]}$						& $10^{-2} (34.8\%) $  \\
                                    &     &     &           &  $ p > 0              $ & $[5,10,20,35,50 \%]$               	& $5\% (39.6\%) $  \\
SVM based on LPCA's extracted features only (\ref{LPCA})										& yes	& no	&SVM-L-A   	&  $ C > 0							$ & $10^{[-6,-4,-2,0,2,4,6]}$  					&	$10^2 (47.3\%) $  \\
                                    &     &     &           &  $ p > 0              $ & $[5,10,20,35,50 \%]$          			& $5\% (69.5\%) $  \\	
SVM based on BoP modularity's extracted features (\ref{BoPMod}) 							& yes	& no	&SVM-BoPM-A &  $\theta > 0					$ & $10^{[-9,-6,-3,0]}$                	& $10^0 (35.2\%) $  \\
																		&			&			&						&  $ C > 0							$ & $10^{[-6,-4,-2,0,2,4,6]}$ 					& $10^3 (44.4\%) $  \\
                                    &     &     &           &  $ p > 0              $ & $[5,10,20,35,50 \%]$          			& $5\% (72.0\%) $  \\
																		\hline
SVM based on node features only (\textbf{baseline})  							& no 	& yes	&SVM-X      &  $ C > 0							$ & $10^{[-6,-4,-2,0,2,4,6]}$ 					& $10^{-2} (27.2\%) $  \\
																		\hline
Spatial autoregressive model (\ref{SAR}) 																& yes	& yes	&SAR-AX     &  $ none							  $ & $-$ 												 				& - \\
SVM on Moran and features on nodes (\ref{Moran})						& yes	& yes	&SVM-M-AX   &  $ C > 0							$ & $10^{[-6,-4,-2,0,2,4,6]}$ 					& $10^2 (26.9\%) $  \\
                                    &     &     &           &  $ p > 0							$ & $[5,10,20,35,50 \%]$            		& $5\% (33.3\%) $  \\
SVM on Geary and features on nodes (\ref{Geary})						& yes	& yes	&SVM-G-AX   &  $ C > 0							$ & $10^{[-6,-4,-2,0,2,4,6]}$ 					&	$10^2 (21.2\%) $  \\
                                    &     &     &           &  $ p > 0							$ & $[5,10,20,35,50 \%]$  						  & $5\% (31.8\%) $  \\
SVM on LPCA and features on nodes (\ref{LPCA})						& yes	& yes	&SVM-L-AX   &  $ C > 0							$ & $10^{[-6,-4,-2,0,2,4,6]}$ 					& $10^2 (28.4\%) $  \\
                                    &     &     &           &  $ p > 0              $ & $[5,10,20,35,50 \%]$               	& $5\% (41.8\%) $  \\
SVM on BoP modularity and features on nodes (\ref{BoPMod})	& yes	& yes	&SVM-BoPM-AX&  $\theta > 0					$ & $10^{[-9,-6,-3,0]}$                	& $10^0 (27.4\%) $  \\
																		&			&			&						&  $ C > 0							$ & $10^{[-6,-4,-2,0,2,4,6]}$ 					& $10^3 (41.0\%) $  \\
                                    &     &     &           &  $ p > 0              $ & $[5,10,20,35,50 \%]$          			& $5\% (48.9\%) $  \\	
SVM on autocovatiates and features on nodes (\ref{auto})	& yes	& yes	&ASVM-AX    &  $ C > 0							$ & $10^{[-6,-4,-2,0,2,4,6]}$ 					&	$10^0 (28.1\%) $  \\
SVM on a double kernel (\ref{DK SVM})								& yes	& yes	&SVM-DK-AX  &  $ C > 0							$ & $10^{[-6,-4,-2,0,2,4,6]}$						&	$10^{-4} (31.7\%) $  \\

\hline
\end{tabular}
\end{adjustbox} 
\label{Param} 
                        
\end{table*}

		\begin{table*}[htp]
\caption{Classification accuracy in percent $\pm$ standard deviation, averaged over the 5 runs, obtained for the 14 methods and the ten datasets. Results are reported for the five feature sets (100F stands for the set of 100 features, and so on). The standard deviation is computed over the 5 folds of the external cross-validation and the 5 independent runs. Bold results are the best obtained performance for each different dataset and feature set.}
\scriptsize
\begin{center}
\begin{tabular}{@{\,\,}c@{} @{\,\,}c@{}|c|cccccccccc}\hline
&& {\textbf{$l$}} & \small{DB1}& \small{DB2}& \small{DB3}& \small{DB4}& \small{DB5}& \small{DB6}& \small{DB7}& \small{DB8} & \small{DB9} & \small{DB10}\\

      \hline 
		\multirow{5}{*}{\rotatebox{90}{SAR}}&
    \multirow{5}{0.3cm}{\rotatebox{90}{AX}}
    &100F&53.18$\pm$12.66&62.90$\pm$5.20&64.16$\pm$12.41&70.55$\pm$7.87&50.68$\pm$25.83&72.71$\pm$23.86&56.36$\pm$24.09&61.18$\pm$9.57&77.50$\pm$1.34&32.07$\pm$10.75\\
    &&50F&66.94$\pm$9.75&66.54$\pm$5.19&65.86$\pm$8.12&71.98$\pm$7.00&66.15$\pm$28.63&78.89$\pm$17.94&61.48$\pm$22.45&64.52$\pm$3.16&64.28$\pm$5.23&35.64$\pm$7.59\\
    &&25F&65.43$\pm$10.55&66.90$\pm$5.20&69.07$\pm$4.48&68.08$\pm$6.99&80.29$\pm$20.51&86.28$\pm$10.30&69.68$\pm$12.28&56.04$\pm$7.31&53.60$\pm$9.30&35.44$\pm$5.74\\
    &&10F&64.37$\pm$5.13&64.84$\pm$5.62&63.56$\pm$4.99&67.70$\pm$6.16&76.14$\pm$9.22&78.32$\pm$9.27&66.02$\pm$13.80&44.18$\pm$10.32&42.93$\pm$9.14&29.03$\pm$2.48\\
    &&5F&62.03$\pm$4.93&55.70$\pm$8.39&60.47$\pm$6.69&61.64$\pm$8.26&74.70$\pm$7.70&78.12$\pm$9.39&66.17$\pm$13.90&42.74$\pm$11.03&37.07$\pm$6.25&21.20$\pm$1.84\\
    \hline 
		\multirow{5}{*}{\rotatebox{90}{{SVM-G}}}&
    \multirow{5}{0.3cm}{\rotatebox{90}{AX}}
    &100F&83.99$\pm$3.19&\textbf{80.98$\pm$2.36}&80.77$\pm$3.84&83.17$\pm$2.00&87.82$\pm$1.76&91.63$\pm$1.72&74.94$\pm$2.05&70.46$\pm$1.06&71.37$\pm$1.54&54.62$\pm$0.91\\
    &&50F&79.51$\pm$3.81&76.64$\pm$2.64&77.21$\pm$3.27&78.47$\pm$3.81&87.59$\pm$1.35&89.65$\pm$1.10&77.45$\pm$1.78&62.92$\pm$1.21&65.98$\pm$1.27&45.94$\pm$2.42\\
    &&25F&69.77$\pm$4.36&70.60$\pm$4.48&73.57$\pm$3.18&74.32$\pm$3.69&90.36$\pm$1.24&91.65$\pm$1.61&79.31$\pm$1.05&59.92$\pm$1.46&67.51$\pm$2.18&43.40$\pm$1.70\\
    &&10F&63.03$\pm$2.62&58.89$\pm$5.68&64.00$\pm$4.36&64.76$\pm$4.32&93.12$\pm$1.61&93.47$\pm$1.27&78.70$\pm$3.10&57.11$\pm$2.15&72.06$\pm$1.77&39.23$\pm$1.22\\
    &&5F&57.31$\pm$4.58&56.83$\pm$6.35&64.28$\pm$5.44&59.56$\pm$3.16&93.21$\pm$2.14&93.34$\pm$1.53&79.86$\pm$2.68&57.14$\pm$2.57&73.19$\pm$2.70&35.90$\pm$1.29\\
    \hline 
		\multirow{5}{*}{\rotatebox{90}{{SVM-M}}}&
    \multirow{5}{0.3cm}{\rotatebox{90}{AX}}
    &100F&83.67$\pm$3.50&80.62$\pm$3.32&80.70$\pm$3.97&83.25$\pm$2.17&87.68$\pm$2.06&93.17$\pm$3.56&74.91$\pm$2.62&70.47$\pm$1.19&71.28$\pm$1.24&54.41$\pm$0.87\\
    &&50F&79.46$\pm$3.72&76.45$\pm$3.08&77.31$\pm$2.97&78.51$\pm$4.85&89.93$\pm$1.19&94.97$\pm$2.92&76.03$\pm$1.78&64.10$\pm$1.57&72.07$\pm$2.60&46.67$\pm$2.00\\
    &&25F&68.10$\pm$5.00&71.01$\pm$4.75&73.05$\pm$3.34&74.42$\pm$3.99&93.73$\pm$2.60&96.56$\pm$2.01&79.94$\pm$2.65&65.41$\pm$1.08&73.83$\pm$1.63&45.50$\pm$1.21\\
    &&10F&62.21$\pm$4.30&57.29$\pm$6.67&63.67$\pm$3.94&64.23$\pm$4.05&94.91$\pm$1.73&97.59$\pm$1.43&80.66$\pm$2.55&66.33$\pm$2.15&76.36$\pm$1.19&41.91$\pm$1.74\\
    &&5F&58.11$\pm$4.65&56.37$\pm$5.53&65.35$\pm$2.69&59.38$\pm$3.04&94.14$\pm$2.01&97.55$\pm$1.57&79.78$\pm$2.72&67.51$\pm$3.51&76.06$\pm$1.01&38.74$\pm$1.77\\
    \hline 
		\multirow{5}{*}{\rotatebox{90}{{BoP}}}&
    \multirow{5}{0.3cm}{\rotatebox{90}{A}}
    &100F&54.45$\pm$3.65&41.84$\pm$4.25&48.42$\pm$0.85&45.70$\pm$0.21&\textbf{96.76$\pm$10.34}&\textbf{98.63$\pm$11.08}&82.48$\pm$6.06&69.91$\pm$19.19&78.11$\pm$19.89&35.34$\pm$4.16\\
    &&50F&54.45$\pm$3.65&41.84$\pm$4.25&48.51$\pm$0.90&45.72$\pm$0.21&\textbf{96.74$\pm$12.06}&\textbf{98.62$\pm$11.82}&82.50$\pm$6.02&69.91$\pm$19.19&78.11$\pm$19.89&35.34$\pm$4.16\\
    &&25F&54.45$\pm$3.65&41.84$\pm$4.25&48.33$\pm$0.88&45.64$\pm$0.15&\textbf{96.80$\pm$11.26}&\textbf{98.62$\pm$10.15}&82.50$\pm$6.04&69.91$\pm$19.19&78.11$\pm$19.89&35.34$\pm$4.16\\
    &&10F&54.45$\pm$3.65&41.84$\pm$4.25&48.33$\pm$0.90&45.70$\pm$0.32&\textbf{96.76$\pm$12.06}&\textbf{98.63$\pm$11.08}&82.49$\pm$6.00&69.91$\pm$19.19&78.11$\pm$19.89&35.34$\pm$4.16\\
    &&5F&54.45$\pm$3.65&41.84$\pm$4.25&48.30$\pm$0.86&45.70$\pm$0.25&\textbf{96.77$\pm$10.33}&\textbf{98.63$\pm$11.82}&82.50$\pm$6.07&69.91$\pm$19.19&78.11$\pm$19.89&35.34$\pm$4.16\\
    \hline 
		\multirow{5}{*}{\rotatebox{90}{{CTK}}}&
    \multirow{5}{0.3cm}{\rotatebox{90}{A}}
    &100F&54.21$\pm$2.66&42.43$\pm$4.06&46.70$\pm$3.13&42.94$\pm$7.17&96.30$\pm$0.33&98.44$\pm$0.40&\textbf{82.87$\pm$1.21}&70.46$\pm$1.27&\textbf{81.69$\pm$0.78}&36.38$\pm$1.00\\
    &&50F&54.21$\pm$2.66&42.43$\pm$4.06&47.25$\pm$2.80&42.88$\pm$7.14&96.31$\pm$0.34&98.44$\pm$0.38&\textbf{82.92$\pm$1.17}&\textbf{70.46$\pm$1.27}&\textbf{81.69$\pm$0.78}&36.38$\pm$1.00\\
    &&25F&54.21$\pm$2.66&42.41$\pm$4.06&46.86$\pm$2.95&42.92$\pm$7.17&96.27$\pm$0.31&98.44$\pm$0.40&\textbf{82.91$\pm$1.18}&\textbf{70.46$\pm$1.27}&\textbf{81.69$\pm$0.78}&36.38$\pm$1.00\\
    &&10F&54.21$\pm$2.66&42.49$\pm$3.95&46.87$\pm$3.02&42.98$\pm$7.17&96.32$\pm$0.34&98.41$\pm$0.41&\textbf{82.92$\pm$1.19}&70.46$\pm$1.27&\textbf{81.69$\pm$0.78}&36.38$\pm$1.00\\
    &&5F&53.86$\pm$3.13&42.46$\pm$3.95&47.14$\pm$2.85&42.94$\pm$7.17&96.30$\pm$0.29&98.42$\pm$0.38&\textbf{82.90$\pm$1.17}&70.46$\pm$1.27&\textbf{81.69$\pm$0.78}&36.38$\pm$1.00\\
    \hline 
		\multirow{5}{*}{\rotatebox{90}{SVM}}&
    \multirow{5}{0.3cm}{\rotatebox{90}{X}}
    &100F&83.72$\pm$3.70&80.76$\pm$2.46&80.93$\pm$3.88&83.11$\pm$2.54&88.45$\pm$1.10&91.41$\pm$1.15&74.66$\pm$2.65&70.46$\pm$1.12&71.41$\pm$0.89&54.59$\pm$1.06\\
    &&50F&80.60$\pm$2.34&\textbf{76.91$\pm$2.78}&\textbf{78.33$\pm$3.24}&81.12$\pm$4.07&89.10$\pm$1.35&91.19$\pm$1.11&77.98$\pm$2.24&68.81$\pm$1.00&68.58$\pm$0.85&40.21$\pm$1.51\\
    &&25F&74.56$\pm$3.29&\textbf{75.49$\pm$4.25}&77.52$\pm$2.89&79.42$\pm$3.19&89.45$\pm$1.68&91.76$\pm$1.12&78.89$\pm$5.87&66.68$\pm$1.07&64.18$\pm$0.63&34.23$\pm$0.96\\
    &&10F&70.85$\pm$3.36&\textbf{74.35$\pm$3.78}&75.64$\pm$4.13&71.71$\pm$3.27&89.52$\pm$1.54&91.97$\pm$0.67&80.52$\pm$0.93&59.45$\pm$1.33&56.27$\pm$1.86&30.82$\pm$0.51\\
    &&5F&65.55$\pm$3.13&66.17$\pm$7.07&71.02$\pm$2.88&75.02$\pm$2.44&87.18$\pm$0.92&92.19$\pm$0.49&78.48$\pm$5.79&53.90$\pm$1.60&42.26$\pm$3.22&25.23$\pm$0.46\\
    \hline 
		\multirow{5}{*}{\rotatebox{90}{{SVM-M}}}&
    \multirow{5}{0.3cm}{\rotatebox{90}{A}}
    &100F&46.26$\pm$6.78&33.13$\pm$5.08&47.03$\pm$7.53&40.03$\pm$4.52&95.43$\pm$1.10&95.07$\pm$1.40&81.60$\pm$1.59&55.93$\pm$1.63&74.51$\pm$1.62&30.82$\pm$1.10\\
    &&50F&46.26$\pm$6.78&33.13$\pm$5.08&47.03$\pm$7.36&40.09$\pm$4.52&95.43$\pm$1.10&95.06$\pm$1.53&81.61$\pm$1.59&55.93$\pm$1.63&74.51$\pm$1.62&30.82$\pm$1.10\\
    &&25F&46.26$\pm$6.78&32.96$\pm$5.08&47.10$\pm$7.51&40.09$\pm$4.52&95.43$\pm$1.09&95.15$\pm$1.53&81.61$\pm$1.59&55.93$\pm$1.63&74.51$\pm$1.62&30.82$\pm$1.10\\
    &&10F&46.29$\pm$6.82&33.13$\pm$5.08&47.17$\pm$7.46&40.03$\pm$4.50&95.43$\pm$1.09&95.08$\pm$1.53&81.60$\pm$1.59&55.93$\pm$1.63&74.51$\pm$1.62&30.82$\pm$1.10\\
    &&5F&46.39$\pm$6.78&33.04$\pm$5.08&47.14$\pm$7.51&40.03$\pm$4.50&95.43$\pm$1.09&95.07$\pm$1.53&81.61$\pm$1.59&55.93$\pm$1.63&74.51$\pm$1.62&30.82$\pm$1.10\\
    \hline 
		\multirow{5}{*}{\rotatebox{90}{{SVM-G}}}&
    \multirow{5}{0.3cm}{\rotatebox{90}{A}}
    &100F&43.45$\pm$8.24&33.32$\pm$6.54&44.10$\pm$4.30&39.97$\pm$4.93&89.82$\pm$12.23&93.79$\pm$1.53&78.53$\pm$1.25&68.13$\pm$1.93&75.58$\pm$1.16&35.03$\pm$1.49\\
    &&50F&43.45$\pm$8.54&33.13$\pm$6.55&44.21$\pm$4.03&39.79$\pm$4.93&89.82$\pm$12.23&93.67$\pm$1.55&78.53$\pm$1.24&68.13$\pm$1.93&75.58$\pm$1.16&35.03$\pm$1.49\\
    &&25F&43.45$\pm$7.89&33.29$\pm$6.54&44.21$\pm$4.10&39.81$\pm$4.48&89.98$\pm$12.20&93.80$\pm$1.54&78.52$\pm$1.24&68.14$\pm$1.93&75.58$\pm$1.16&35.03$\pm$1.49\\
    &&10F&43.45$\pm$7.89&33.15$\pm$6.60&44.21$\pm$4.23&39.97$\pm$4.93&89.95$\pm$12.23&93.54$\pm$1.44&78.52$\pm$1.25&68.13$\pm$1.93&75.58$\pm$1.16&35.03$\pm$1.49\\
    &&5F&43.45$\pm$8.54&33.15$\pm$6.57&44.17$\pm$4.03&39.81$\pm$4.93&89.83$\pm$12.23&93.66$\pm$1.53&78.52$\pm$1.24&68.13$\pm$1.93&75.58$\pm$1.16&35.03$\pm$1.49\\
    \hline 
		\multirow{5}{*}{\rotatebox{90}{{ASVM}}}&
    \multirow{5}{0.3cm}{\rotatebox{90}{AX}}
    &100F&79.38$\pm$3.71&75.08$\pm$3.78&80.10$\pm$3.29&81.12$\pm$3.69&92.07$\pm$1.74&92.65$\pm$2.27&79.80$\pm$1.17&66.14$\pm$1.97&70.58$\pm$3.58&44.55$\pm$4.77\\
    &&50F&79.21$\pm$4.09&74.40$\pm$2.84&77.66$\pm$3.51&79.41$\pm$4.43&92.32$\pm$2.16&94.76$\pm$1.90&80.50$\pm$2.41&66.08$\pm$1.63&76.28$\pm$2.43&37.13$\pm$7.96\\
    &&25F&\textbf{74.64$\pm$4.89}&73.27$\pm$3.50&75.74$\pm$3.76&\textbf{79.80$\pm$3.48}&95.34$\pm$1.80&95.98$\pm$1.52&81.59$\pm$2.75&65.96$\pm$2.59&76.95$\pm$2.46&32.66$\pm$8.29\\
    &&10F&\textbf{71.41$\pm$3.45}&72.29$\pm$6.41&\textbf{75.74$\pm$7.78}&\textbf{76.15$\pm$4.31}&95.48$\pm$1.76&96.65$\pm$1.34&81.16$\pm$2.99&62.93$\pm$3.41&74.93$\pm$2.85&28.24$\pm$6.85\\
    &&5F&\textbf{65.82$\pm$7.35}&\textbf{70.92$\pm$6.94}&69.39$\pm$6.70&\textbf{75.46$\pm$7.53}&95.58$\pm$1.05&96.60$\pm$1.93&80.85$\pm$5.92&61.89$\pm$3.59&71.80$\pm$4.94&25.10$\pm$6.08\\
    \hline 
		\multirow{5}{*}{\rotatebox{90}{{SVM-DK}}}&
    \multirow{5}{0.3cm}{\rotatebox{90}{{AX}}}
    &100F&\textbf{84.89$\pm$2.59}&79.89$\pm$3.40&\textbf{81.31$\pm$3.31}&\textbf{84.34$\pm$2.50}&88.75$\pm$2.04&92.14$\pm$0.91&76.09$\pm$1.95&70.38$\pm$0.76&71.32$\pm$1.20&54.23$\pm$0.88\\
    &&50F&\textbf{80.88$\pm$4.78}&76.79$\pm$2.72&77.74$\pm$5.01&80.05$\pm$4.56&88.88$\pm$3.96&91.54$\pm$3.52&79.62$\pm$3.46&68.50$\pm$2.06&69.43$\pm$3.88&41.90$\pm$2.39\\
    &&25F&73.59$\pm$4.12&74.24$\pm$3.09&76.89$\pm$4.23&78.65$\pm$3.88&89.08$\pm$2.64&91.95$\pm$5.71&80.32$\pm$3.95&70.25$\pm$2.10&74.04$\pm$10.60&36.74$\pm$3.48\\
    &&10F&69.68$\pm$3.54&72.36$\pm$5.40&74.91$\pm$7.21&72.23$\pm$4.44&89.40$\pm$7.05&89.70$\pm$9.53&80.22$\pm$4.34&\textbf{72.71$\pm$1.65}&76.62$\pm$17.59&31.83$\pm$3.17\\
    &&5F&65.17$\pm$5.68&68.11$\pm$8.81&\textbf{71.12$\pm$7.52}&74.42$\pm$9.03&87.25$\pm$8.24&88.54$\pm$17.01&80.48$\pm$4.20&\textbf{71.97$\pm$2.90}&78.01$\pm$19.64&25.82$\pm$2.82\\
    \hline 
		\multirow{5}{*}{\rotatebox{90}{{SVM-BoPM}}}&
    \multirow{5}{0.3cm}{\rotatebox{90}{{AX}}}
    &100F&84.07$\pm$4.00&80.43$\pm$3.02&81.19$\pm$2.98&83.39$\pm$2.48&88.15$\pm$1.54&92.54$\pm$3.27&74.10$\pm$2.15&\textbf{70.60$\pm$0.86}&71.70$\pm$2.17&\textbf{56.24$\pm$2.61}\\
    &&50F&79.05$\pm$4.53&76.58$\pm$2.37&78.31$\pm$6.22&\textbf{81.64$\pm$4.71}&88.94$\pm$2.53&92.92$\pm$2.71&77.75$\pm$3.44&68.67$\pm$3.13&71.94$\pm$2.99&44.97$\pm$2.41\\
    &&25F&73.89$\pm$5.29&73.38$\pm$6.51&\textbf{78.19$\pm$6.81}&79.19$\pm$5.48&91.08$\pm$3.46&94.67$\pm$2.02&79.13$\pm$3.19&66.05$\pm$3.69&76.21$\pm$3.24&41.60$\pm$3.04\\
    &&10F&69.84$\pm$6.26&72.09$\pm$8.59&74.86$\pm$7.14&70.15$\pm$5.75&92.92$\pm$2.05&95.60$\pm$1.81&79.33$\pm$3.21&63.64$\pm$2.17&77.26$\pm$1.85&42.20$\pm$2.55\\
    &&5F&60.38$\pm$7.24&65.70$\pm$9.05&68.45$\pm$9.61&73.70$\pm$9.29&90.13$\pm$2.62&95.00$\pm$2.61&77.12$\pm$7.46&63.05$\pm$3.33&80.35$\pm$1.67&\textbf{41.99$\pm$2.23}\\
    \hline 
		\multirow{5}{*}{\rotatebox{90}{{SVM-L}}}&
    \multirow{5}{0.3cm}{\rotatebox{90}{AX}}
    &100F&83.67$\pm$3.72&80.41$\pm$3.00&80.79$\pm$4.02&83.15$\pm$2.50&87.90$\pm$1.80&91.82$\pm$2.44&75.26$\pm$2.11&70.48$\pm$1.11&71.18$\pm$1.11&54.87$\pm$0.87\\
    &&50F&79.46$\pm$3.62&76.70$\pm$3.02&77.26$\pm$2.97&78.75$\pm$3.75&88.78$\pm$3.47&93.09$\pm$2.43&78.40$\pm$1.96&62.81$\pm$4.09&73.04$\pm$3.04&\textbf{48.56$\pm$1.62}\\
    &&25F&68.57$\pm$6.27&70.99$\pm$4.87&73.45$\pm$4.07&74.11$\pm$5.22&90.77$\pm$2.29&95.20$\pm$3.24&79.95$\pm$1.57&62.33$\pm$2.68&73.92$\pm$3.37&\textbf{46.22$\pm$2.66}\\
    &&10F&62.71$\pm$5.48&58.59$\pm$9.74&64.61$\pm$8.36&65.69$\pm$7.10&93.68$\pm$1.95&96.65$\pm$1.56&81.18$\pm$1.90&61.35$\pm$3.82&76.79$\pm$2.00&\textbf{42.31$\pm$1.98}\\
    &&5F&57.95$\pm$5.99&57.27$\pm$7.30&64.12$\pm$8.45&58.83$\pm$8.49&94.19$\pm$2.30&96.47$\pm$2.80&80.94$\pm$1.80&61.91$\pm$4.95&76.33$\pm$1.99&39.62$\pm$2.44\\
    \hline 
		\multirow{5}{*}{\rotatebox{90}{{SVM-L}}}&
    \multirow{5}{0.3cm}{\rotatebox{90}{A}}
    &100F&40.85$\pm$4.08&33.07$\pm$5.31&40.43$\pm$4.70&42.01$\pm$3.00&90.93$\pm$1.92&95.85$\pm$1.75&79.07$\pm$2.33&62.40$\pm$1.67&76.58$\pm$0.93&34.42$\pm$1.00\\
    &&50F&40.85$\pm$4.10&33.59$\pm$4.88&39.66$\pm$5.11&41.99$\pm$3.00&90.93$\pm$1.92&95.83$\pm$1.75&79.07$\pm$2.33&62.40$\pm$1.67&76.58$\pm$0.93&34.42$\pm$1.00\\
    &&25F&40.74$\pm$4.08&33.96$\pm$5.57&39.56$\pm$5.13&42.01$\pm$3.00&90.96$\pm$1.87&95.80$\pm$1.78&79.07$\pm$2.33&62.40$\pm$1.67&76.58$\pm$0.93&34.42$\pm$1.00\\
    &&10F&40.79$\pm$4.10&33.67$\pm$4.90&39.73$\pm$4.94&42.01$\pm$3.00&90.93$\pm$1.92&95.72$\pm$1.80&79.07$\pm$2.33&62.40$\pm$1.67&76.58$\pm$0.93&34.42$\pm$1.00\\
    &&5F&40.79$\pm$4.13&33.96$\pm$5.57&39.68$\pm$5.11&42.01$\pm$3.00&90.96$\pm$1.87&95.71$\pm$1.80&79.07$\pm$2.33&62.40$\pm$1.67&76.58$\pm$0.93&34.42$\pm$1.00\\
		\hline 
		\multirow{5}{*}{\rotatebox{90}{{SVM-BoPM}}}&
    \multirow{5}{0.3cm}{\rotatebox{90}{{A}}}
    &100F&43.54$\pm$0.78&32.32$\pm$3.38&39.35$\pm$2.34&42.61$\pm$3.32&91.75$\pm$0.63&94.41$\pm$1.12&79.13$\pm$0.87&67.50$\pm$1.16&80.28$\pm$0.31&14.87$\pm$0.34\\
    &&50F&43.54$\pm$0.78&32.32$\pm$3.38&39.35$\pm$2.34&42.61$\pm$3.32&91.75$\pm$0.63&94.41$\pm$1.12&79.13$\pm$0.87&67.50$\pm$1.16&80.28$\pm$0.31&14.87$\pm$0.34\\
    &&25F&43.54$\pm$0.78&32.32$\pm$3.38&39.35$\pm$2.34&42.61$\pm$3.32&91.75$\pm$0.63&94.41$\pm$1.12&79.13$\pm$0.87&67.50$\pm$1.16&80.28$\pm$0.31&14.87$\pm$0.34\\
    &&10F&43.54$\pm$0.78&32.32$\pm$3.38&39.35$\pm$2.34&42.61$\pm$3.32&91.75$\pm$0.63&94.41$\pm$1.12&79.13$\pm$0.87&67.50$\pm$1.16&80.28$\pm$0.31&14.87$\pm$0.34\\
    &&5F&43.54$\pm$0.78&32.32$\pm$3.38&39.35$\pm$2.34&42.61$\pm$3.32&91.75$\pm$0.63&94.41$\pm$1.12&79.13$\pm$0.87&67.50$\pm$1.16&80.28$\pm$0.31&14.87$\pm$0.34\\
\end{tabular}
\label{RES}
\end{center}
\end{table*}

	\begin{figure}[t!]
	\begin{center}
		\includegraphics[scale = 0.66]{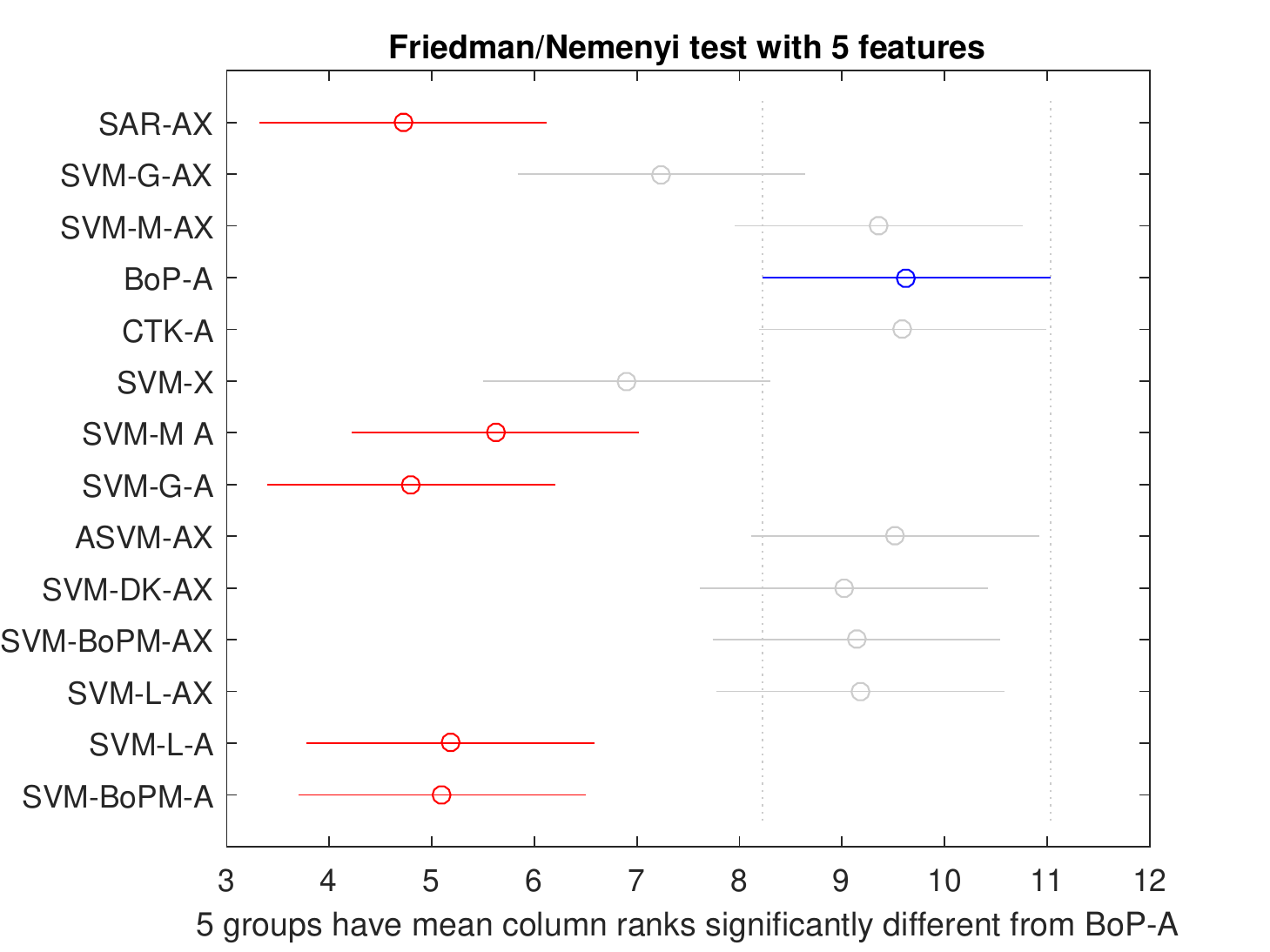}
		\caption{Mean rank (circles) and critical difference (plain line) of the Friedman/Nemenyi test, over 5 runs and all datasets, obtained on partially labeled graphs. The blue method has the best mean rank and is statistically better than red methods. Labeling rate is 20\% and the critical difference is 2.81. This figure shows the results when only 5 node features are considered (5F datasets).}
		\label{FN5}
		\end{center}
		\end{figure}	
		
		
		\begin{figure}[t!]
	\begin{center}
		\includegraphics[scale = 0.66]{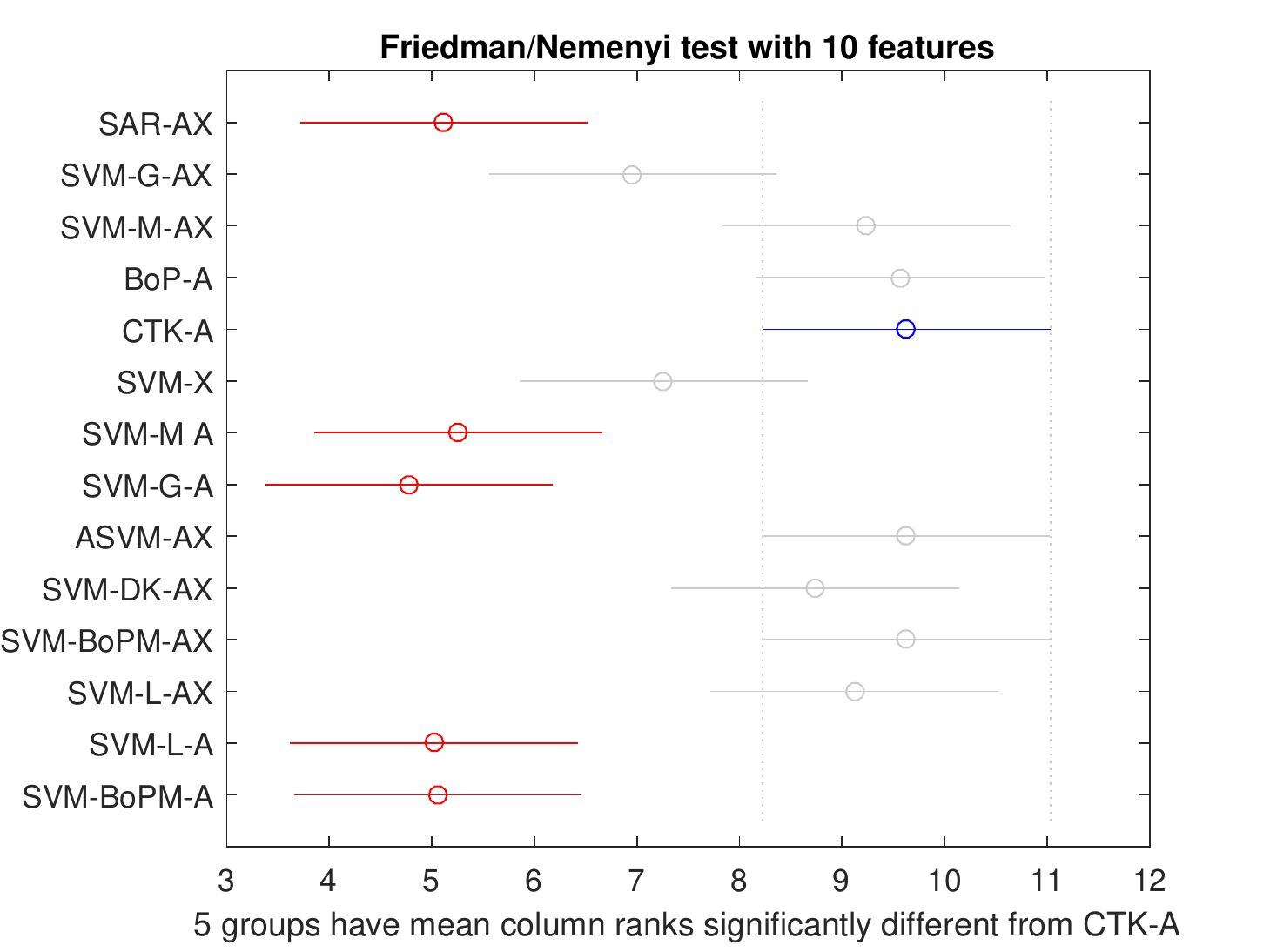}
		\caption{The Friedman/Nemenyi test considering 10 node features (10F datasets); see Figure \ref{FN5} for details. The critical difference is 2.81.}
		\label{FN10}
		\end{center}
		\end{figure}
		
		
		\begin{figure}[t!]
	\begin{center}
		\includegraphics[scale = 0.66]{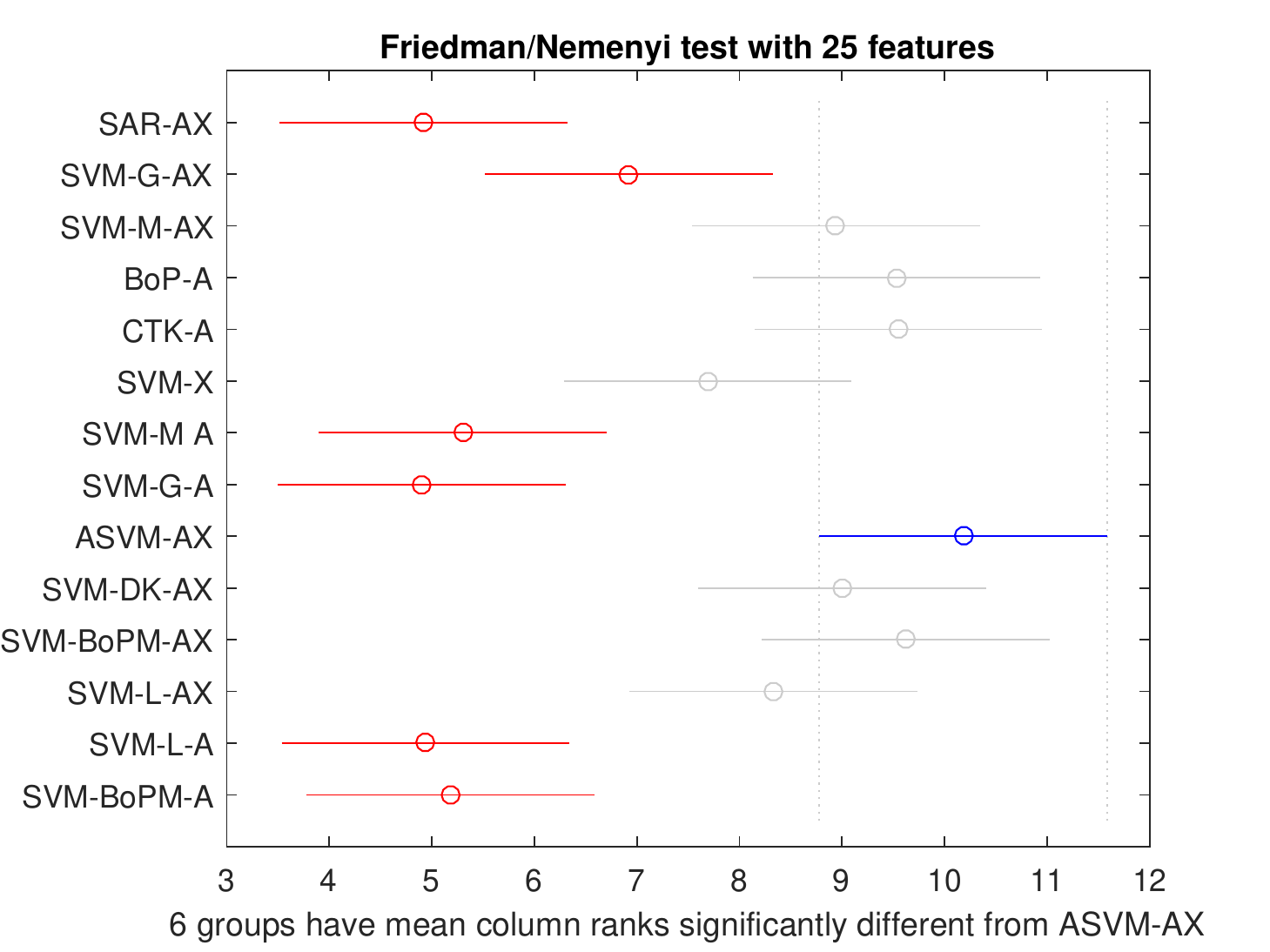}
		\caption{The Friedman/Nemenyi test considering 25 node features (25F datasets); see Figure \ref{FN5} for details. The critical difference is 2.81.}
		\label{FN25}
		\end{center}
		\end{figure}
		
		
		\begin{figure}[t!]
	\begin{center}
		\includegraphics[scale = 0.66]{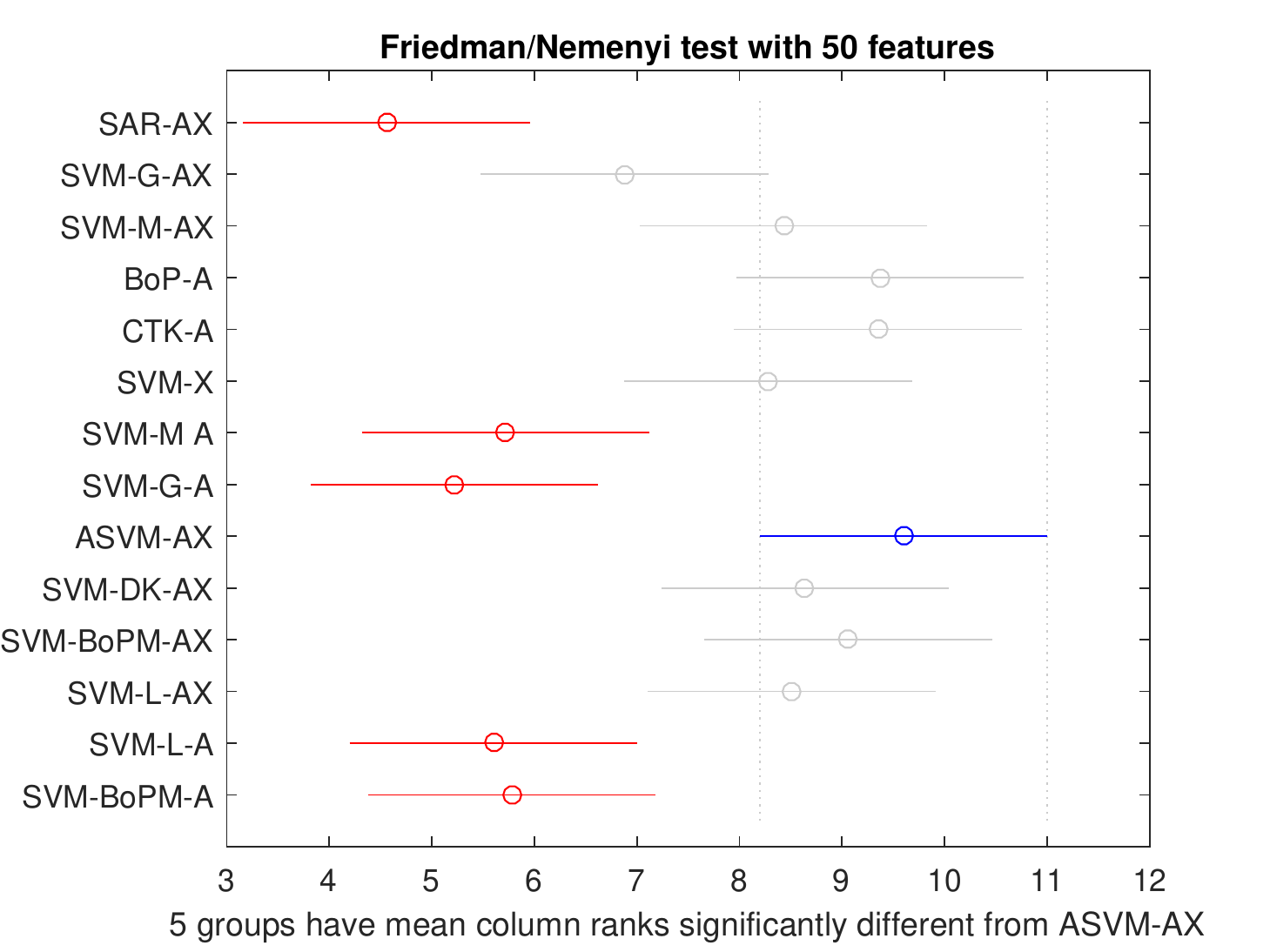}
		\caption{The Friedman/Nemenyi test considering 50 node features (50F datasets); see Figure \ref{FN5} for details. The critical difference is 2.81.}
		\label{FN50}
		\end{center}
		\end{figure}
		
		
		\begin{figure}[t!]
	\begin{center}
		\includegraphics[scale = 0.66]{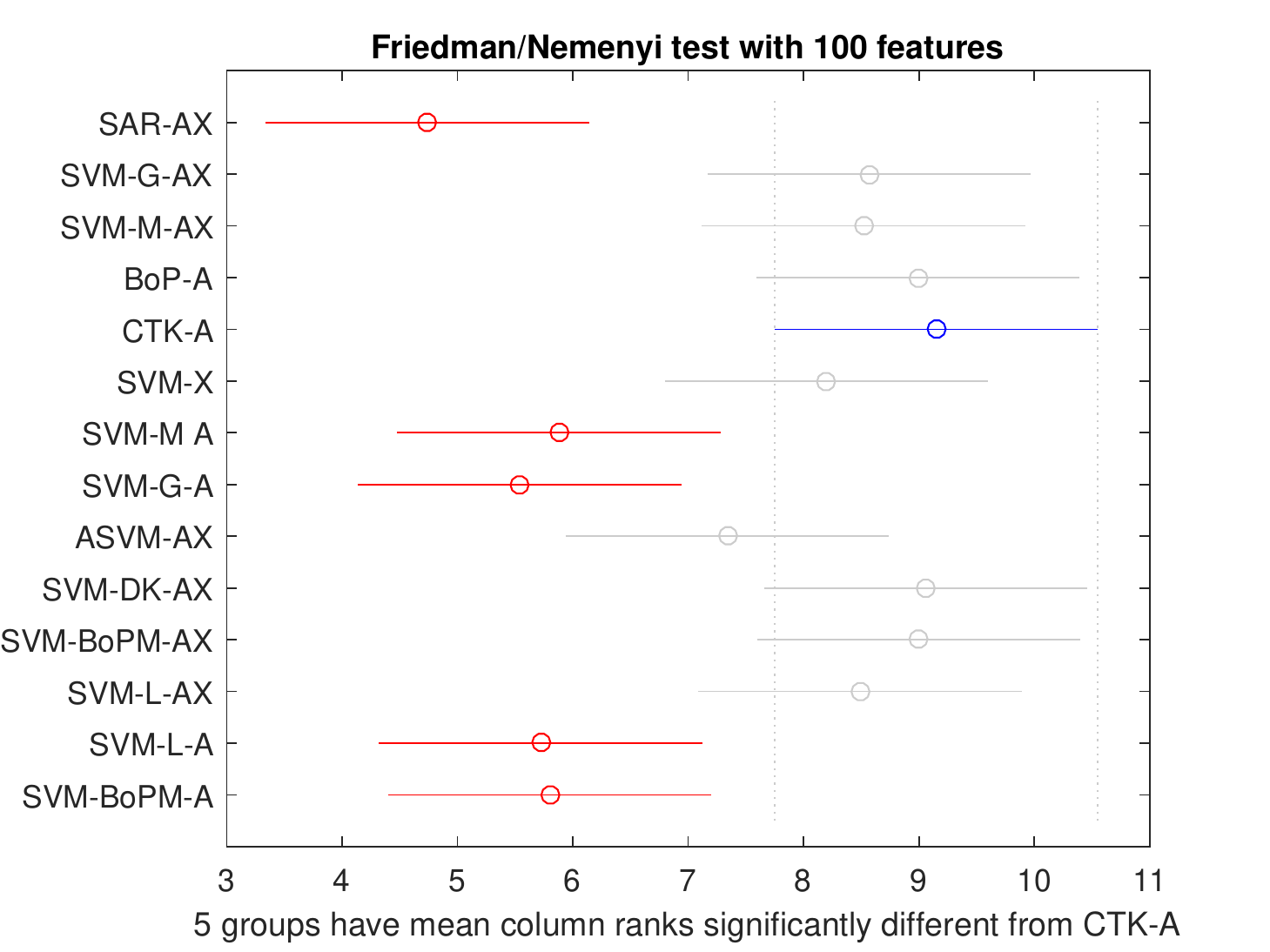}
\caption{The Friedman/Nemenyi test considering 100 node features (100F datasets); see Figure \ref{FN5} for details. The critical difference is 2.81.}
		\label{FN100}
		\end{center}
		\end{figure}
		
		
					\begin{figure}[t!] 
	\begin{center}
		\includegraphics[scale = 0.66]{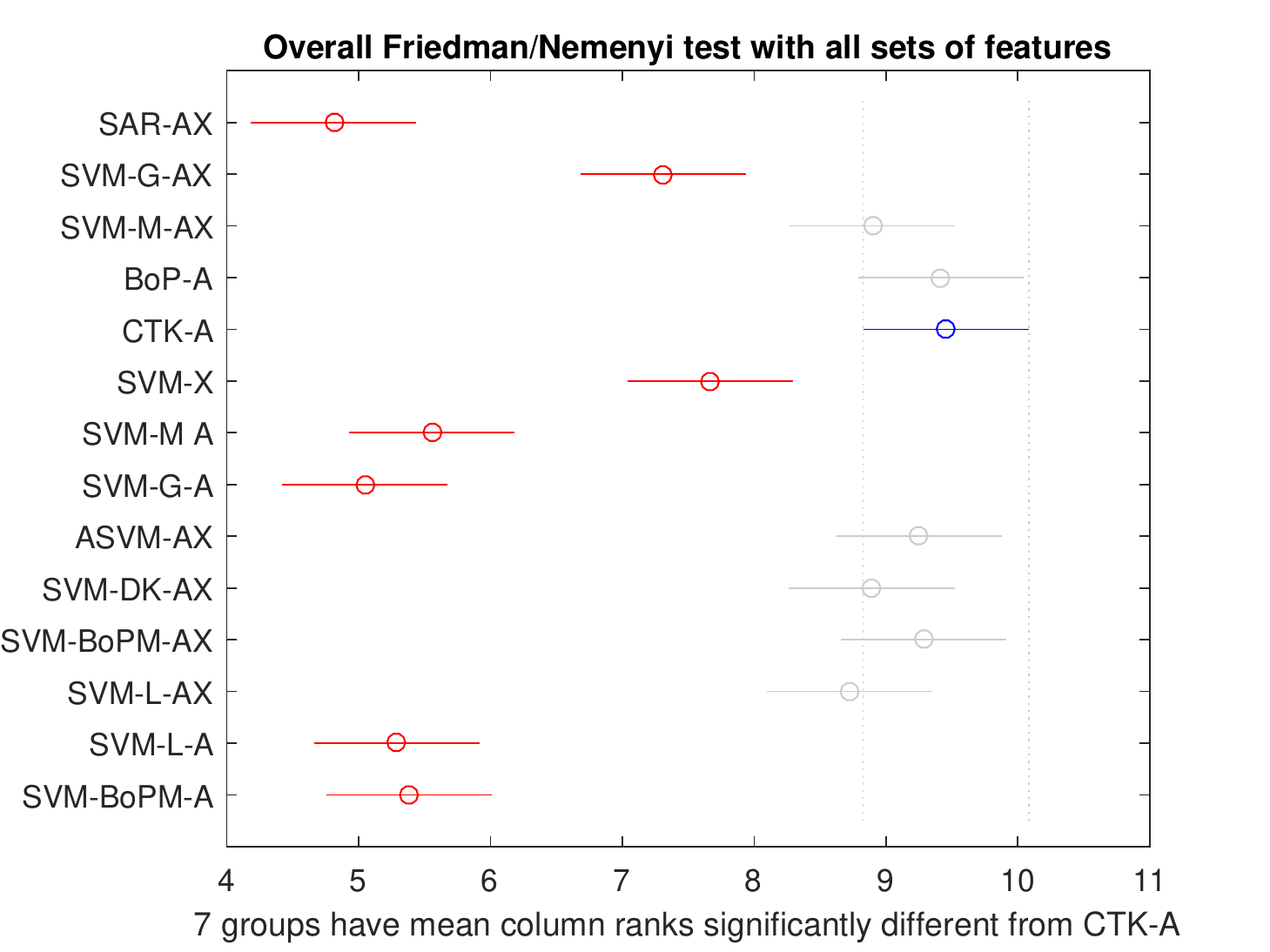}
		\caption{Friedman/Nemenyi test considering all feature sets (5F, 10F, 25F , 50F, 100F). The critical difference is 1.25; see Figure \ref{FN5} for details.}
		\label{FNOV}
		\end{center}
		\end{figure}


	\begin{figure}[t!]
	\begin{center}
		\includegraphics[scale = 0.66]{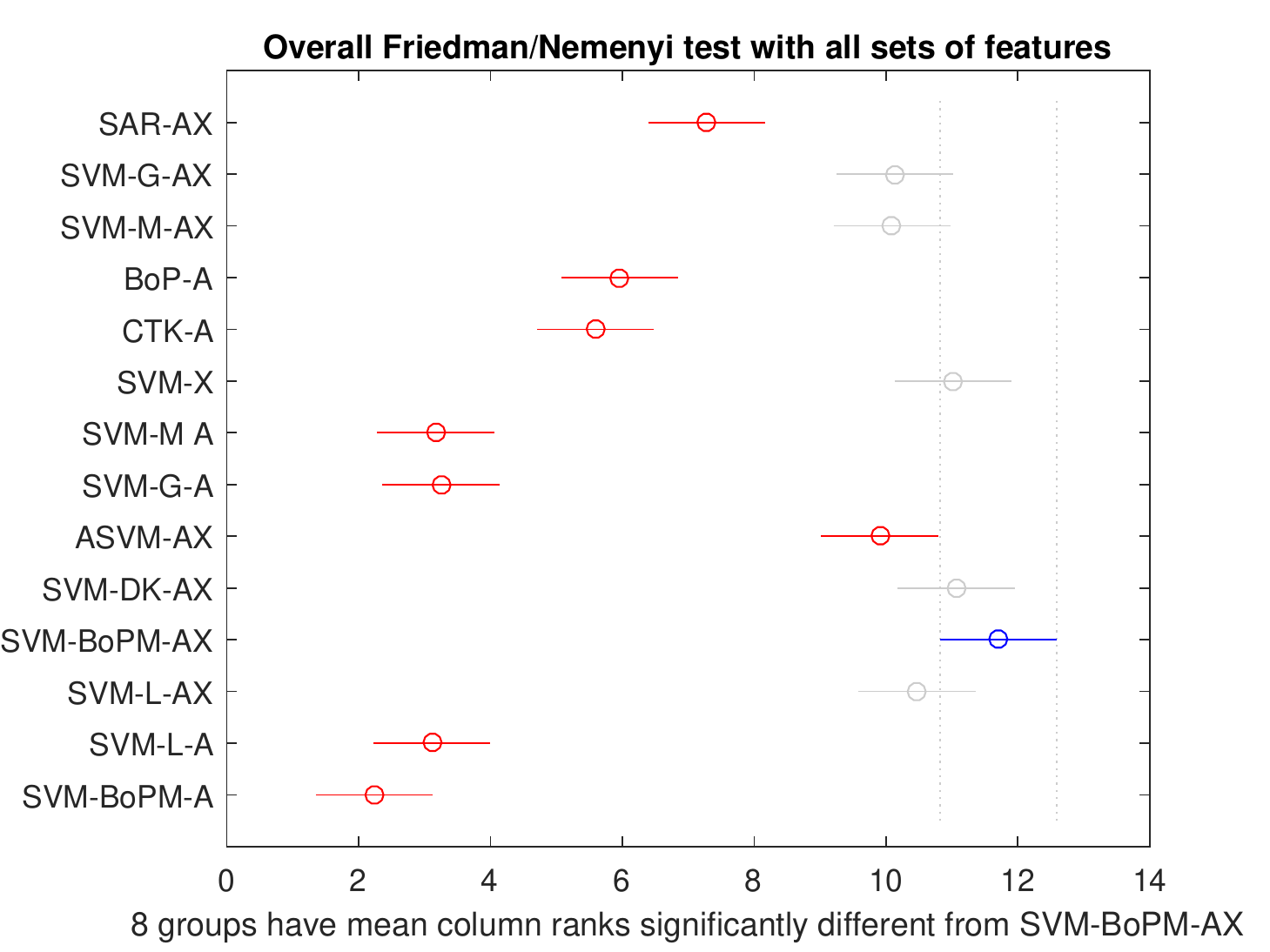}
		\caption{Friedman/Nemenyi test considering all feature sets (5F, 10F, 25F, 50F, 100F), but only on databases DB1 to DB4 and DB10 (driven by node features, $\mathbf{X}$); see Figure \ref{FN5} for details. The critical difference is 1.77.}
		\label{Xdriven}
		\end{center}
		\end{figure}


	\begin{figure}[t!]
	\begin{center}
		\includegraphics[scale = 0.66]{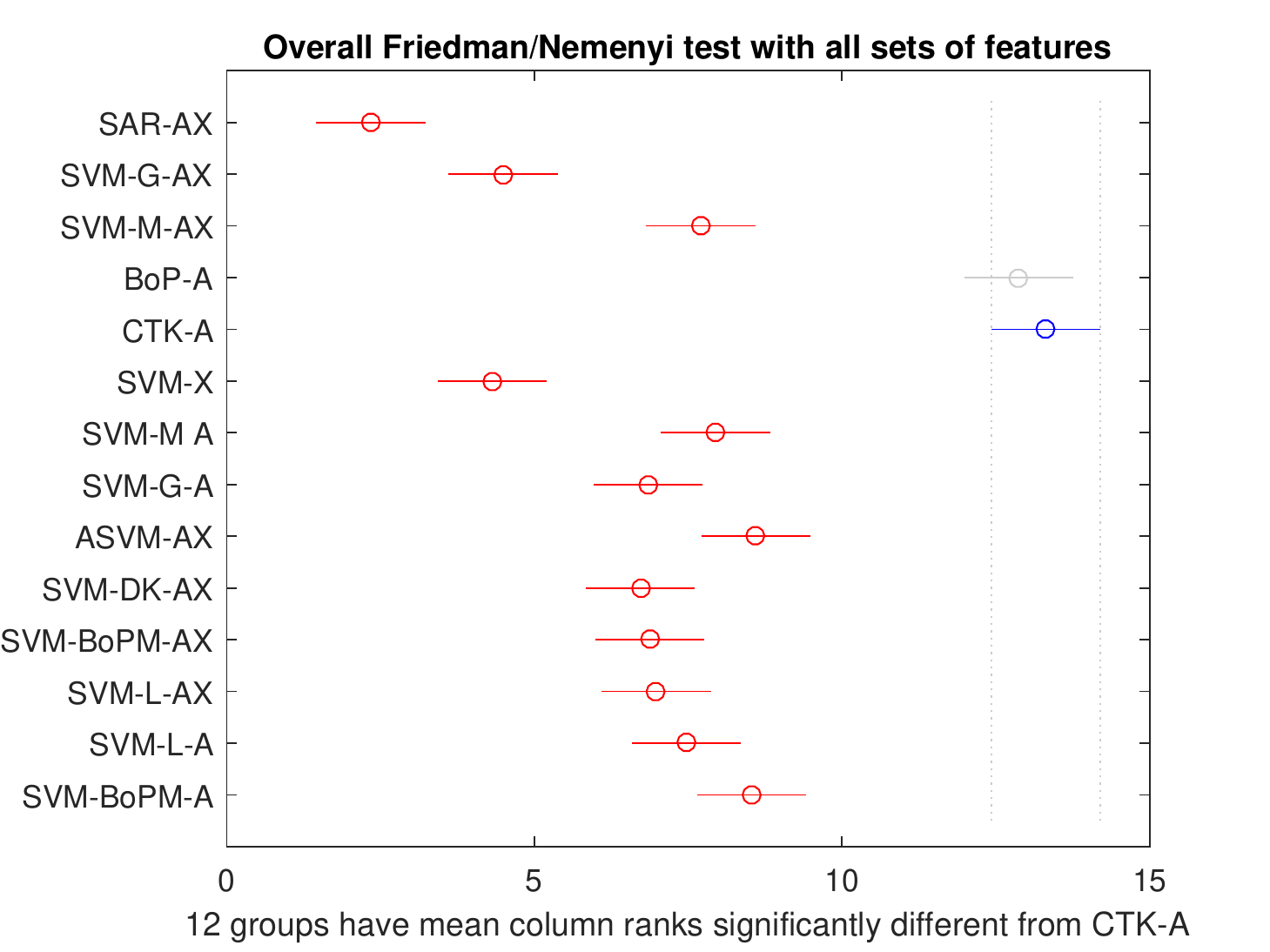}
		\caption{Friedman/Nemenyi test considering all feature sets (5F, 10F, 25F, 50F, 100F), but only on databases DB5 to DB9 (driven by graph structure, $\mathbf{A}$); see Figure \ref{FN5} for details. The critical difference is 1.77.}
		\label{Adriven}
		\end{center}
		\end{figure}


	\begin{figure}[t!] 
	\begin{center}
		\includegraphics[scale = 0.66]{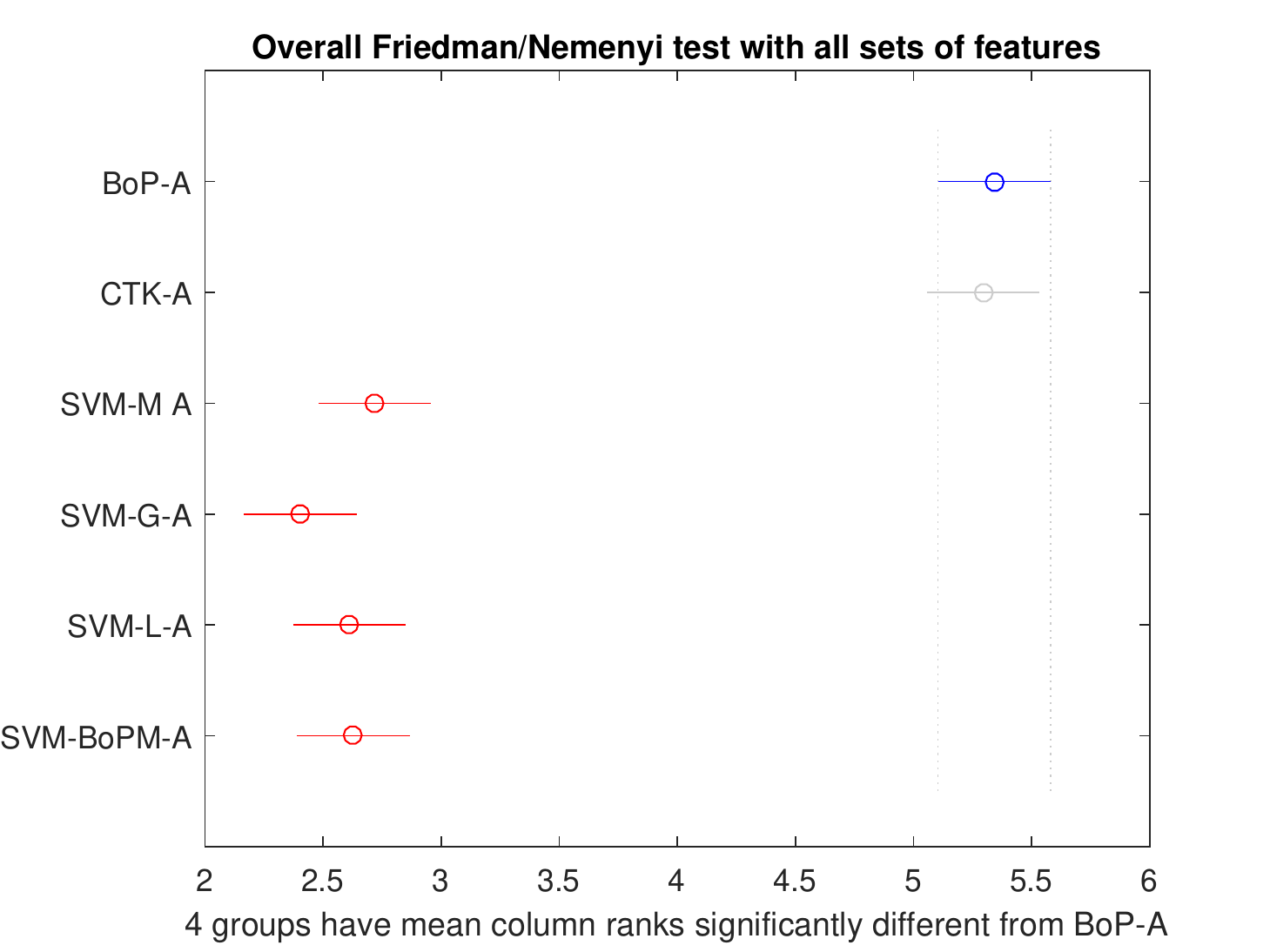}
		\caption{Friedman/Nemenyi test considering all feature sets (5F, 10F, 25F, 50F, 100F), for methods based on graph information alone (A); see Figure \ref{FN5} for details. The critical difference is 0.48.}
		\label{Aonly}
		\end{center}
		\end{figure}


	\begin{figure}[t!] 
	\begin{center}
		\includegraphics[scale = 0.66]{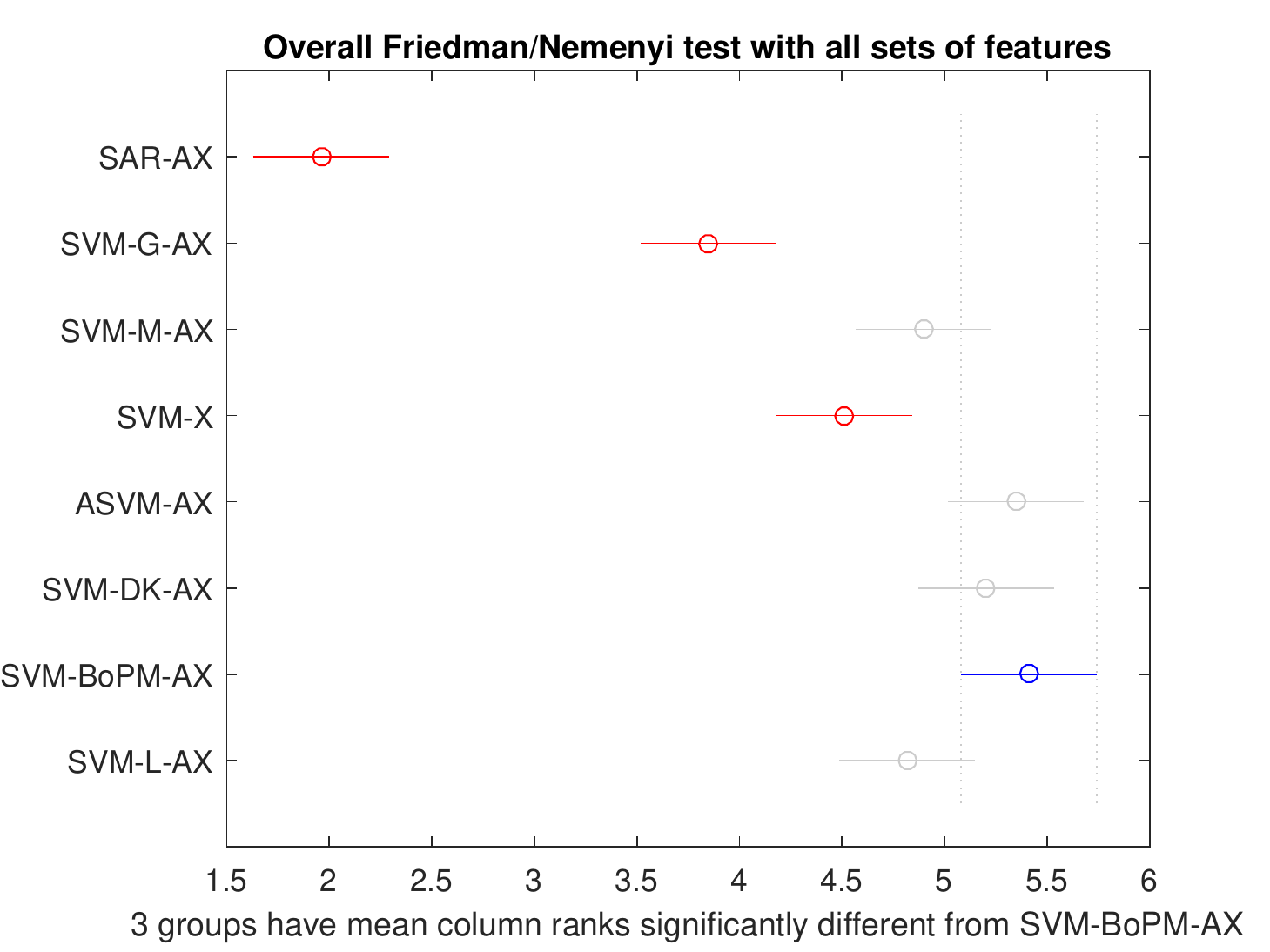}
		\caption{Friedman/Nemenyi test considering all feature sets (5F, 10F, 25F, 50F, 100F), performed only on methods combining graph information and node information (AX, plus simple SVM-X as baseline). See Figure \ref{FN5} for details. The critical difference is 0.66.}
		\label{AXonly}
		\end{center}
		\end{figure}


	\begin{figure}[t!]
	\begin{center}
		\includegraphics[scale = 0.66]{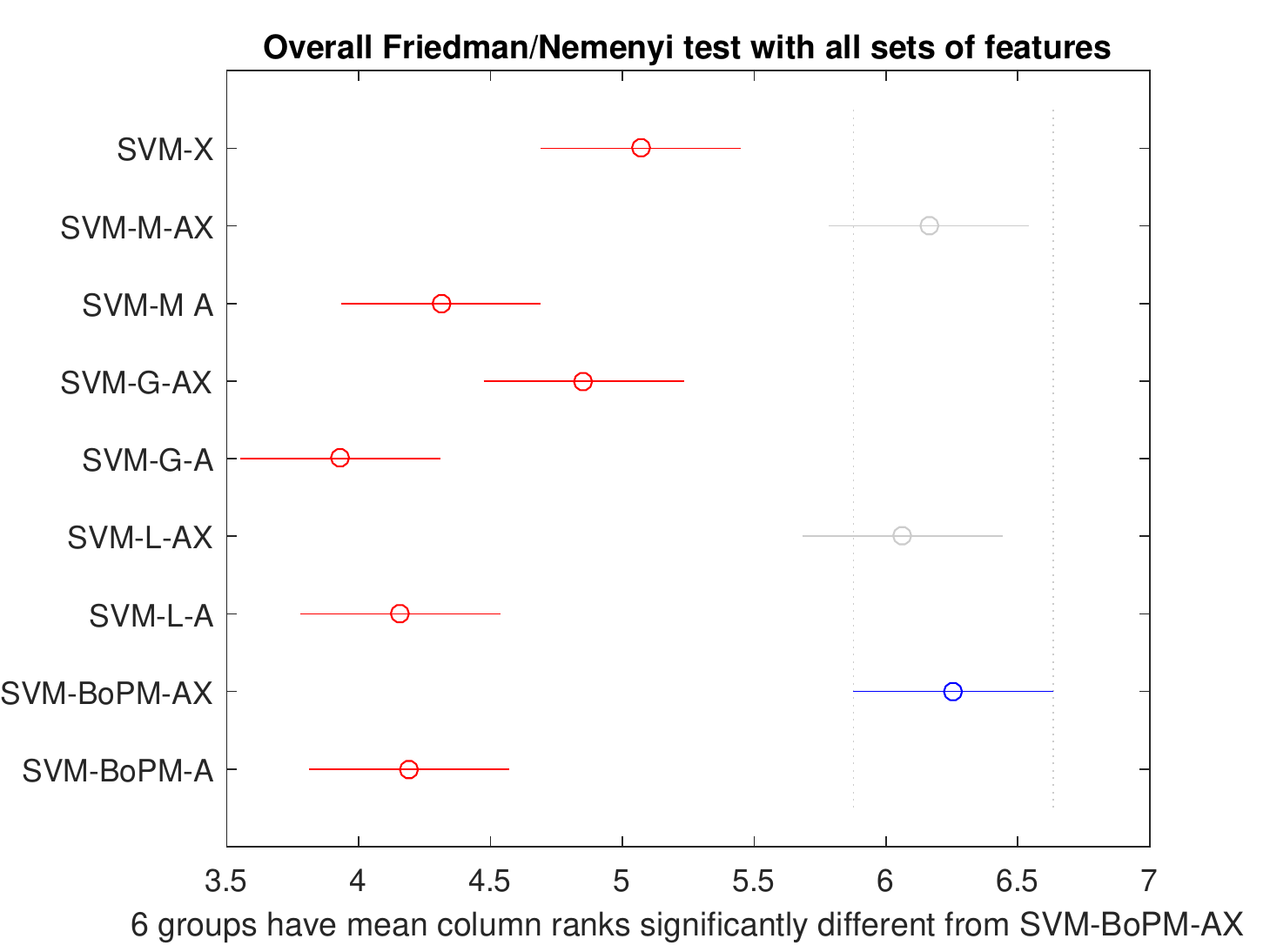}
		\caption{Friedman/Nemenyi test considering all feature sets (5F, 10F, 25F, 50F, 100F), for methods based on a graph embedding (plus regular linear SVM for comparison). See Figure \ref{FN5} for details. The critical difference is 0.76.}
		\label{AvsAX}
		\end{center}
		\end{figure}

\subsection{Experimental methodology}
\label{ExpM}

The classification accuracy will be reported for a 20\% labeling rate i.e. proportion of nodes for which labels are known. Labels of remaining nodes are deleted during model fitting phase and are used as test data during the assessment phase, where the various classification models predict the most suitable category of each unlabeled node in the test set.

For each considered feature sets and for each dataset, samples are randomly assigned into 25 folds: 5 external folds are defined and, for each of them, 5 nested folds are also defined. This procedure is performed 5 time to get 5 \emph{runs} for each feature sets and dataset.
The performances for one specific run are then computed by taking the average over the 5 external folds and, for each of them, a 5-fold nested cross-validation is performed to tune the parameters of the models on the inner folds (see Table \ref{Param}).

\subsection{Results and discussion}
\label{ResDis}

First of all, most frequently selected parameters values are indicated on Table \ref{Param}. We observe that the most selected value for $p$ (the number of eigenvectors extracted for representing the graph structure; see Section \ref{DimR}) is actually low. This is a good news since efficient eigensystem solvers can be used to compute the first eigenvectors corresponding to the largest (or smallest) eigenvalues.

The classification accuracy and standard deviation, averaged on the 5 runs, are reported on Table \ref{RES}, for the 10 different datasets and the 5 sets of features.
Bold values indicate the best performance on those 50 combinations. Recall that the BoP-A, CTK-A, SVM-M-A, SVM-G-A, and SVM-L-A methods do not depend on the node features as they are based on the graph structure only. It explains why results do not depend on feature set for those five methods. 

Moreover, the different classifiers have been compared across datasets through a Friedman test and a Nemenyi post-hoc test \cite{Demsar2006}. The Friedman test is a non-parametric equivalent of the repeated-measures ANOVA. It ranks the methods for each dataset separately, the best algorithm getting the rank 1, the second best rank 2, etc. Once the null hypothesis (the mean ranking of all methods is equal, meaning all classifiers are equivalent) is rejected with $p$-value $< 0.05$, the (non parametric) post-hoc Nemenyi test is then computed. Notice that all Friedman tests were found to be positive. The Nemenyi test determines whether or not each method is significantly better ($p$-value less than 0.05 based on the 5 runs) to another. This is reported, for each feature set in turn (5F, 10F, \dots, 100F), and thus increasing information available on the nodes, in Figures \ref{FN5} to \ref{FN100}, and an overall test based on all the features sets and datasets is reported in Figure \ref{FNOV}.

\subsubsection{Overall performances on all datasets and all node feature sets}

From Table \ref{RES} and Figure \ref{FNOV}, overall best performances on all dataset and all node feature sets are often obtained either by a SVM based on node features combined with new features derived from the graph structure (Subsection \ref{DimR}), or, unexpectedly, by the CTK-A sum-of-similarities method (using graph structure only; see Subsection \ref{RCTK}), which performs quite well on datasets five to nine.
The BoP-A node betweenness (using graph structure only, see Subsection \ref{BOP}) is also competitive and achieves results similar to the sum-of-similarities CTK-A method (as already observed in \cite{MOI}).

On the contrary, the best method among the graph structure plus node features SVM is not straightforward do determine (see Figure \ref{FNOV}). From Figures \ref{FN5} to \ref{FN100}, the main trend is that the performance decreases when the number of features decreases, which seems normal.

However, this trend is not always observed; for example, with the SVM-M-AX method (SVM with features extracted from Moran's index and features on nodes, see Subsection \ref{Moran}) and database DB5, the performances rise when the number of features decreases. This can be explained if we realize that each dataset can be better described in terms of its graph structure (graph-driven dataset, databases DB5 to DB9), or by its node features (features-driven dataset, databases DB1 to DB4 and DB10).

To highlight this behavior, the network autocorrelation for each class (i.e., for each $\mathbf{y}^{c}$) was computed and the average is reported for each dataset. This measure quantifies to which extent the target variable is correlated through on neighboring nodes. The values are reported on Table \ref{MGLPCA} for Moran's $I$ , Geary's $c$ and LPCA contiguity ratio (see Subsection \ref{Moran}). For Moran's $I$, high values (large autocorrelation) indicates that the graph structure is highly informative. This is the opposite for Geary and LPCA, as small values correspond to large autocorrelation.

Nevertheless, from Table \ref{RES} and Figure \ref{FNOV}, the best overall performing methods combining node features and graph structure are (excluding the methods based on the graph alone, BoP-A and CTK-A), SVM-BoPM-AX (SVM with bag-of-paths modularity, see Subsection \ref{BoPMod}) and ASVM-AX (SVM based on autocovariates, see Subsection \ref{auto}). They are not statistically different from SVM-M-AX, SVM-L-AX and SVM-DK-AX, but their mean rank is slightly higher.

Notice also that, from Figure \ref{FNOV}, if we look at the performances obtained by a baseline linear SVM based on node features only (SVM-X), we clearly observe that integrating the information extracted from the graph structure significantly improves the results. Therefore, it seems to be be a good idea to consider collecting link information which could improve the predictions.

\begin{table}
\caption{Mean autocorrelation of class memberships. For Moran's $I$, high value implies large autocorrelation. For Geary's $c$ and LPCA, small value implies large autocorrelation. For each measure, \textbf{-} indicates the bound of maximum negative autocorrelation, a value close to \textbf{0} indicates a lack of structural association and \textbf{+} indicates the bound of maximum positive autocorrelation. See Subsection \ref{DimR} for details. Datasets can be divided into two groups (more driven by $\mathbf{A}$ or by $\mathbf{X}$) according to these measures.}
\scriptsize
\begin{center}
\begin{tabular}{|l|c|c|c|c|c|c|c|c|}
\hline
\textbf{$\mathbf{A}$-driven} &\textbf{+}& \textbf{0} & \textbf{-} & DB 1		&	DB 2		&	DB 3		&	DB 4		&	DB 10		\\
\hline	
Moran's $I$ 	& -1 & 0 & 1 & 1.27  & 1.09  & 0.66  & 0.53  & 0.79	\\
Geary's $c$ 	&  0 & 1 & 2 & 0.09  & 0.09  & 0.33  & 0.19  & 0.12	\\
\cline{3-4}
LPCA c. ratio & ~0\footnotemark[1]{} & \multicolumn{2}{c|}{$>1$ \footnotemark[1]{}} & 0.20  & 0.13  & 0.58  & 0.67  & 0.26	\\
\hline
\textbf{$\mathbf{X}$-driven} &\textbf{+}& \textbf{0} & \textbf{-} & DB 5		&	DB 6  	&	DB 7		&	DB 8		&	DB 9		\\
\hline
Moran's $I$ 	& -1 & 0 & 1 & -0.22 &-0.12  &-0.15  &-0.06  & 0.15	\\
Geary's $c$ 	&  0 & 1 & 2 & 0.78  & 0.59  & 0.63  & 0.57  & 0.43	\\
\cline{3-4}
LPCA c. ratio & 0\footnotemark[1]{} & \multicolumn{2}{c|}{$>1$ \footnotemark[1]{}} & 2.54  & 2.10  & 1.86  & 1.90  & 0.82	\\
\hline
\end{tabular}
\label{MGLPCA}
\end{center}
\end{table}

\footnotetext[1]{LPCA continuity ration is positive and lower-bounded by $1-\sqrt{\lambda_{max}}$ (which tends to be close to zero) where $\lambda_{max}$ is the largest eigenvalue of $\mathbf{A}$, the upper bound is unknown \cite{Lebart-2000}.}

\subsubsection{Exploiting either the structure of the graph or the node features alone}

Obviously, as already mentioned, databases DB5 to DB9 are graph-driven, which explains the good performances of the sum-of-similarities CTK-A and BoP-A on these data. For these datasets, the features on the nodes do not help much for predicting the class label, as observed when looking to Figure \ref{Adriven} where results are displayed only on these datasets. It also explains the behavior of method SVM-M-AX on database DB5, among others.

In this case, the best performing methods are the sum-of-similarities CTK-A and the bag-of-paths betweenness BoP-A (see Subsection \ref{GBC}). This is clearly confirmed by displaying the results of the methods based on the graph structure only in Figure \ref{Aonly} and the results obtained on the graph-driven datasets in Figure \ref{Adriven}. Interestingly, in this setting, these two methods ignoring the node features (CTK-A and BoP-A) are outperforming the SVM-based methods.

Conversely, on the node features-driven datasets (DB1 to DB4 and DB10; results displayed in Figure \ref{Xdriven} and Table \ref{RES}), all SVM methods based on node features and graph structure perform well while the methods based on the graph structure only obtain much worse results, as expected.
In this setting, the two best performing methods are the same as for the overall results, that is, SVM-BoPM-AX (SVM with bag-of-paths modularity, see Subsection \ref{BoPMod}) and SVM-DK-AX (SVM based on a double kernel, see Subsection \ref{DK SVM}). However, these methods are not significantly better than a simple linear SVM based on features only (SVM-X), as shown in Figure \ref{Xdriven}. 

From another point of view, Figure \ref{AXonly} takes into account all datasets and compares only the methods combining node features and graph structure. In this setting, the two best methods are also SVM-BoPM-AX, ASVM-AX and SVM-DK-AX which are now significantly better than the baseline SVM-X (less methods and more datasets are compared).

\subsubsection{Comparison of graph embedding methods}

Concerning the embedding methods described in Subsection \ref{DimR}, we can conclude that Geary's index (SVM-G-A and SVM-G-AX) should be avoided by preferring the bag-of-paths modularity (SVM-BoP-AX), Moran's index (SVM-M-AX) or graph Local Principal Component Analysis (SVM-L-AX). This is clearly observable when displaying only the results of the methods combining node features and graph structure in Figure \ref{AvsAX}.
This result is confirmed when comparing only the methods based on graph embedding in Figure \ref{Aonly}.

\subsubsection{Main findings}

To summarize, the experiments lead to the following conclusions:
\begin{itemize}
\item The best performing methods are highly dependent on the dataset. We observed (see Table \ref{MGLPCA}) that, quite naturally, some datasets are more graph-driven in the sense that the network structure conveys important information for predicting the class labels, while other datasets are node features-driven and, in this case, the graph structure does not help much. However, it is always a good idea to consider information about the graph structure as this additional information can improve significantly the results (see Figures \ref{AXonly} and \ref{AvsAX}).
\item If we consider the graph structure alone, the two best investigated methods are the sum-of-similarities CTK-A and the bag-of-paths betweenness BoP-A (see Subsection \ref{GBC}). They clearly outperform the graph embedding methods, but also the SVMs on some datasets.
\item When informative features on nodes are available, it is always a good idea to combine the information, and we found that the best performing methods are SVM-BoPM-AX (SVM with bag-of-paths modularity, see Subsection \ref{BoPMod}), ASVM-AX (SVM based on autocovariates, see Subsection \ref{auto}) and SVM-DK-AX (SVM based on a double kernel, see Subsection \ref{DK SVM}) (see Figure \ref{AXonly}). Taking the graph structure into account clearly improves the results over a baseline SVM based on node features only. 
\end{itemize}

\section{Conclusion}
\label{CCL}

This work considered a data structure made of a graph and plain features on nodes of the graph. 14 semi-supervised classification methods were investigated to compare the feature-based approach, the graph structure-based approach, and the dual approach combining both information sources. 
It appears that the best performance are often obtained either by a SVM method (the considered baseline classifier) based on plain node features combined to a given number of new features derived from the graph structure (namely from the BoP modularity or autocovariates), or by the sum-of-similarities and the bag-of-paths modularity method, based on the graph structure only, which perform well on some datasets for which the graph structure carries important class information.

Indeed, we observed empirically that some datasets can be better explained by their graph structure (graph-driven datasets), or by their node features (features-driven datasets). Consequently, neither the graph-derived features alone or the plain features alone is sufficient to obtain optimal performances. In other words, standard feature-based classification results can be improved significantly by integrating information from the graph structure. In particular, the most effective methods were based on bag-of-paths modularity (SVM-BoPM-AX), autocovariates (ASVM-AX) or a double kernel (SVM-DK-AX).

The take-away message can be summarize as follow: if the dataset is graph-driven, a simple sum-of-similarities or a bag-of-paths betweenness are sufficient but it is not the case if the features on the nodes are more informative. In both cases SVM-BoPM-AX, ASVM-AX, SVM-DK-AX still ensure good overall performances, as shown on the investigated datasets. 

A key point is therefore to determine \emph{a-priori} if a given dataset is graph-driven or features-driven. In this paper we proposed spatial autocorrelation indexes to tackle this issue. Further investigations will be carried in that direction. In particular, how can we automatically infer properties of a new dataset (graph-driven or features-driven) if all class labels are not known (during cross-validation for example)?

Finally, the present work does not take into account the scalability and the complexity point of view. This analysis is left for further work.

\section*{Acknowledgement}{Acknowledgement: This work was partially supported by the Elis-IT project funded by the ``R\'{e}gion wallonne" and the Brufence project supported by INNOVIRIS (``R\'{e}gion bruxelloise"). We thank this institution for giving us the opportunity to conduct both fundamental and applied research.}

\bibliographystyle{elsarticle-num} 
\bibliography{Biblio}




\end{document}